%% file: ArXiv.tex
\documentclass[10pt,twocolumn,letterpaper]{article}

\usepackage[pagenumbers]{cvpr} 

\input{preamble}

\definecolor{cvprblue}{rgb}{0.21,0.49,0.74}
\usepackage[pagebackref,breaklinks,colorlinks,allcolors=cvprblue]{hyperref}

\def\paperID{45684} 
\def\confName{CVPR}
\def\confYear{2026}

\title{Training-Only Heterogeneous Image-Patch-Text Graph Supervision for Advancing Few-Shot Learning Adapters}

\author{
Mohammed Rahman Sherif Khan Mohammad \quad Ardhendu Behera\textsuperscript{\Letter} \quad Sandip Pradhan\\
Swagat Kumar \quad Amr Ahmed\\
Edge Hill University, Ormskirk, Lancashire, L39 4QP\\
{\tt\small \{Mohamm, Beheraa, Pradhans, Kumars, Ahmeda\}@edgehill.ac.uk}
}

\begin{document}
\maketitle
\input{sec/0_abstract}    
\input{sec/1_intro}
\input{sec/2_related_works}
\input{sec/3_method}

\input{sec/4_experiments}
\input{sec/5_conclusion}

{
    \small
    \bibliographystyle{ieeenat_fullname}
    \bibliography{main}
}
\input{sec/X_suppl}


\end{document}

%% file: preamble.tex
\newcommand{\TODO}[1]{\textbf{\color{red}[TODO: #1]}}
\renewcommand{\TODO}[1]{}

\usepackage{microtype}

\usepackage{graphicx}
\usepackage{amsmath}
\usepackage{amssymb}
\usepackage{booktabs}

\usepackage{multirow}
\usepackage{amsfonts} 
\usepackage[table]{xcolor}
\usepackage{subcaption}
\usepackage{svg}
\usepackage{pifont}
\usepackage{marvosym}
\usepackage{tabularx}
\usepackage{float}

\usepackage[accsupp]{axessibility}
\usepackage{caption}
\definecolor{rowhighlight}{gray}{0.92}

\newcommand{\gmark}{\textcolor{green!60!black}{\ding{51}}} 
\newcommand{\rmark}{\textcolor{red}{\ding{55}}}

%% file: sec/0_abstract.tex
\begin{abstract}
Recent adapter-based CLIP tuning (e.g., Tip-Adapter) is a strong few-shot learner, achieving efficiency by caching support features for fast prototype matching. However, these methods rely on global uni-modal feature vectors, overlooking fine-grained patch relations and their structural alignment with class text. To bridge this gap without incurring inference costs, we introduce a novel asymmetric training-only framework. Instead of altering the lightweight adapter, we construct a high-capacity auxiliary Heterogeneous Graph Teacher that operates solely during training. This teacher (i) integrates multi-scale visual patches and text prompts into a unified graph, (ii) performs deep cross-modal reasoning via a Modality-aware Graph Transformer (MGT), and (iii) applies discriminative node filtering to extract high-fidelity class features. Crucially, we employ a cache-aware dual-objective strategy to supervise this relational knowledge directly into the Tip-Adapter’s key–value cache, effectively upgrading the prototypes while the graph teacher is discarded at test time. Thus, inference remains identical to Tip-Adapter with zero extra latency or memory. Across standard 1-16-shot benchmarks, our method consistently establishes a new state-of-the-art. Ablations confirm that the auxiliary graph supervision, text-guided reasoning, and node filtering are the essential ingredients for robust few-shot adaptation. Code is available at \url{https://github.com/MR-Sherif/TOGA.git}.
\end{abstract}

%% file: sec/1_intro.tex
\section{Introduction}
\label{sec:intro}

\begingroup
\renewcommand\thefootnote{}
\footnotetext{\textsuperscript{\Letter} Corresponding author: Ardhendu Behera (Beheraa@edgehill.ac.uk)}
\endgroup

\begin{figure}[!t]
\centering
\includegraphics[width=0.95\linewidth]{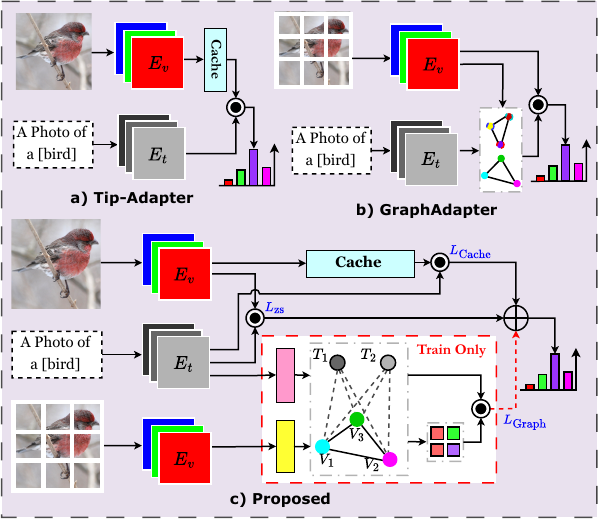}
\caption{The core trade-off of VLM adaptation and our proposed solution.
(a) \emph{Lightweight adapters} (e.g., Tip-Adapter \cite{tipadapter}) are fast but limited, as they only reason over the single, global image feature.
(b) \emph{Heavyweight adapters} (e.g., GraphAdapter \cite{li2023graphadapter}) are more powerful, reasoning over visual patches, but their complex GNN adds a permanent, significant cost at inference time.
(c) \emph{Our proposed method}, \textbf{TOGA} (the best of both), uses a powerful auxiliary teacher. This teacher, our modality-aware graph transformer (MGT), performs deep bi-modal reasoning on a unified graph of patches and text. It supervises its knowledge into the lightweight ``student'' adapter, which is the only component used at inference, achieving high performance with the efficiency of (a).}
\label{fig:teaser}
\end{figure}

Foundational vision-language models (VLMs) such as CLIP \cite{radford2021clip}, ALIGN \cite{jia2021scaling} and LiT \cite{zhai2022lit} achieve impressive zero-shot generalization by aligning images and text in a shared embedding space. Although effective for general-domain recognition, their reliance on a single spatially-averaged global image descriptor is sub-optimal for Fine-Grained Visual Classification (FGVC) \cite{wah2011caltech, krause20133d, fu2017look, he2022transfg}, where sub-categories differ by subtle, localized cues (e.g., beak shape or plumage pattern distinguishing a ``Yellow-bellied Warbler'' from a ``Grey Warbler''). Spatial averaging obscures fine-grained information. Adapting VLMs to few-shot domains remains a key open challenge: naively fine-tuning all parameters is both computationally expensive and prone to overfitting in the $K$-shot regime \cite{tipadapter, zhou2022learning}.

Parameter-Efficient Fine-Tuning (PEFT) offers a pragmatic alternative, but forces a performance-latency trade-off (Fig.~\ref{fig:teaser}). \textbf{Global-feature adapters} (e.g., CLIP-Adapter, Tip-Adapter in Fig. \ref{fig:teaser}a)~\cite{houlsby2019parameter, gao2024clipadapter, tipadapter} are attractive for their tiny parameter footprint and fast inference (CLIP-Adapter adds near-constant overhead; Tip-Adapter’s cost scales with cache size). However, they still operate on \textit{semantically diluted} global features, leaving fine-grained evidence under-exploited. 
\textbf{Patch-level adapters} (e.g., Visual Prompt Tuning (VPT)~\cite{jia2022visual}, GraphAdapter \cite{li2023graphadapter} in Fig.~\ref{fig:teaser}b) enrich representations by reasoning over patch tokens, but at the price of permanent inference overhead (extra tokens, GNN passes) and often unimodal reasoning (patch$\leftrightarrow$patch) that under-utilizes text semantics. In summary, current PEFT is either fast but coarse or expressive but slow at inference.

\noindent\textbf{Motivation:} This exposes a clear research gap. The ideal few-shot adapter must satisfy two competing objectives: (1) be expressive enough to reason over fine-grained, patch-level evidence and its \textit{bimodal} alignment to textual class semantics; (2) preserve the zero-overhead inference of lightweight baselines (e.g., Tip-Adapter). Modern VLM adapters do not achieve both: global-feature adapters are fast but coarse (they inherit the spatially averaged descriptor), while patch-level adapters improve detail sensitivity but introduce GNN/extra-token computation at test time and are often \textit{unimodal} (patch$\leftrightarrow$patch only).

Our approach is simple: inject relational cross-modal reasoning only during training and guide the components that remain at deployment: the adapter and its key–value cache. We introduce \textbf{TOGA} (Training-Only Graph Adapter), an asymmetric distillation framework (Fig. \ref{fig:teaser}(c)) co-training a lightweight, permanent student (cache adapter) with a powerful, training-only teacher. The novelty lies in the teacher's architecture, which is a modality-aware \textit{Heterogeneous Image-Patch–Text Graph} built from frozen CLIP encoders. It represents multi-scale visual patches and class-name text embeddings as nodes within a single graph topology. A Modality-aware Graph Transformer (MGT) performs type-specific message passing over image$\leftrightarrow$patch, patch$\leftrightarrow$patch, image$\leftrightarrow$text and patch$\leftrightarrow$text edges to learn the complex, fine-grained correlations between visual evidence and class semantics. This knowledge is dynamically distilled into the student adapter (key-value cache) through cache-aware objectives. At testing, the entire teacher stack is discarded: we train with graphs, test as Tip-Adapter with no architectural changes, no extra latency or memory. 

This is effective because (i) the teacher injects high-level relational knowledge (disambiguating near-neighbors via patch co-occurrence/consistency) and cross-modal selectivity (pulling evidence toward the correct prompt, suppressing noise). (ii) Distillation targets the key-value cache mechanism to maintain gains without the teacher at the test time.
Our contributions are: (1) A novel asymmetric training-only distillation for few-shot VLM adaptation that couples the Tip-Adapter key-value cache (student) with a high-capacity teacher, delivering zero test-time overhead. (2) A Modality-aware Graph Transformer over a unified heterogeneous image-patch–text graph that performs deep bi-modal (visual and text) hierarchical (image$\leftrightarrow$patch) reasoning; to our knowledge, this is the first such teacher for few-shot CLIP adaptation. (3) We introduce a cache-aware dual-objective co-training strategy where a Focal Loss \cite{lin2017focal} is used as a teacher-forcing regularizer to ensure that the auxiliary graph teacher becomes a robust expert. (4) Comprehensive evidence on 1–16-shot benchmarks: consistent gains over lightweight (Tip-/CLIP-Adapter) and heavyweight (graph/patch) baselines, with ablations isolating the effects of patch-only graphs, patch–text edges, focalized teacher forcing, and cache alignment.

%% file: sec/2_related_works.tex
\section{Related Works}
\label{sec:related-works}

\textbf{Parameter-Efficient VLM Adaptation.}
Adapting large VLMs without full fine-tuning is costly and overfit-prone in a few-shot regimes \cite{tipadapter, zhou2022learning}. This has led to three dominant PEFT strands: 

(1) \textit{Prompt learning}: CoOp~\cite{zhou2022learning} introduces learnable textual context tokens for CLIP, but its static prompts overfit base classes and generalize poorly to unseen categories. CoCoOp~\cite{zhou2022conditional} addresses this with input-conditional prompts that adapt per image, improving base$\rightarrow$novel transfer. KgCoOp~\cite{yao2023visual} further regularizes prompt tuning by pulling learned prompts toward hand-crafted prompts, mitigating the forgetting of CLIP’s general knowledge. ProGrad~\cite{zhu2023prompt} constrains optimization via prompt-aligned gradients to prevent catastrophic drift during tuning. MaPLe~\cite{khattak2023maple} couples visual and textual prompts and learns stage-wise prompts to tighten cross-modal alignment. PLOT++~\cite{chen2022prompt} learns multiple prompts and uses optimal transport to match prompt distributions to visual features, avoiding prompt collapse. Similarly, Visual Fourier Prompt Tuning \cite{zeng2024visual} and semantic-guided DA-VPT \cite{ren2025vpt} learn prompts in visual space. Despite these advances, most prompt methods still reason over a global image descriptor, not patch-level structures, and therefore have limited leverage on fine-grained cues.

(2) \textit{Adapters and residual heads}: 
This strand adds lightweight modules, such as small bottleneck layers (CLIP-Adapter) \cite{gao2024clipadapter} or low-rank updates (LoRA) \cite{hu2022lora, zanella2024low}, to the frozen encoders. TaskRes~\cite{yu2023task} acts directly on the text-based classifier through a residual that decouples prior vs. task knowledge (good preservation of CLIP priors). Recent task-aware variants (Ta-Adapter~\cite{zhang2024ta}) equip visual and textual encoders with task prompts. Yet, like prompt methods, these operate largely at the global feature level and do not inject structured patch-wise reasoning.

(3) \textit{Cache-based methods}: 
Tip-Adapter~\cite{tipadapter} constructs a key–value cache over few-shot supports for fast prototype matching and can be lightly fine-tuned; follow-ups explore cache design, calibration, and hybrid cascades (e.g., CaFo~\cite{zhang2023prompt}: prompt$\rightarrow$generate$\rightarrow$cache, COSMIC~\cite{huang2025cosmic}, or other two-stage adaptation strategies~\cite{farina2025rethinking}).
These lines keep inference cheap but again rely on global embeddings, under-utilizing relational patch evidence. Our work directly supervises the cache—keys/values—via training-only relational signals.

\noindent\textbf{Graph Neural Networks (GNNs) for Visual Reasoning.} A complementary line model relations and spatial structure at patch/region level using GNNs. In the context of FGVC, GNNs capture higher-order visual interactions \cite{sikdar2025interweaving} and spatial structure \cite{bera2022srgnn}. GraphAdapter \cite{li2023graphadapter} builds dual knowledge graphs (visual and textual) for VLM tuning, improving alignment but still requires graph processing at test time (and is not cache-centric). In broader multimodal tasks (Visual Question Answering (VQA), Scene Graph Generation (SGG)), unified heterogeneous graphs directly connect visual and textual nodes constructed from pixels \cite{li2024pixels} or 3D objects \cite{wu20243d}, allowing cross-modal message passing. However, these designs are rarely used as training-only teachers for few-shot VLM adaptation. Our approach fills this gap: a heterogeneous image-patch–text graph runs only during training, supervising the adapter to encode fine-grained knowledge into its key–value cache.

\noindent\textbf{Knowledge Distillation and Auxiliary Training.} Classical KD transfers soft logits from a teacher to a student; in VLMs, most works use offline teachers or symmetric mutual learning. TaskRes~\cite{yu2023task} can be viewed as preserving priors by a residual head; ProGrad~\cite{zhu2023prompt} KgCoOp~\cite{yao2023visual} can be seen as regularized supervision that guards general knowledge. We adopt an asymmetric online scheme in which a high-capacity teacher (heterogeneous image-patch–text graph) supervises a lightweight student (cache adapter) and goes beyond logits by directly supervising cache keys/values, directly upgrading the prototype memory that governs the test-time behavior of the Tip-Adapter. This yields the expressivity of graph reasoning without carrying its cost to deployment.

%% file: sec/3_method.tex
\section{Proposed Method}
\label{sec:method}
Our framework (Fig.~\ref{fig:architecture}) comprises three components: (1) a frozen CLIP zero-shot branch (ZS) \cite{radford2021clip}; (2) a lightweight student adapter, i.e., the key–value cache model of Tip-Adapter-F \cite{tipadapter}; and (3) \textbf{TOGA}, a heterogeneous image-patch–text graph teacher (highlighted by the {\color{red}dotted red} rectangle). The core novelty is an asymmetric supervision scheme: the student adapter is optimized under the guidance of the high-capacity graph teacher, which injects fine-grained, cross-modal relational cues during training. In inference, the teacher is entirely discarded. This implies that the deployment path, latency, and memory remain the same as in the Tip-Adapter-F.
\subsection{Problem Formulation: Zero-Shot/Tip-Adapter}
\label{sec:baseline}
Given a $K$-shot dataset with $C$ classes and $N=K\times C$ labeled supports, we adapt a frozen CLIP model with image encoder $f_v(.)\in \mathbb{R}^D$ and text encoder $f_t(.)\in \mathbb{R}^D$, where $D$ is the embedding dimension. Let $T=\{f_t(t_c)\}_{c=1}^C$ be class prompt embeddings and $x$ is a query image with global feature $z=f_v(x)$. The zero-shot logit $L_{ZS}=\text{cos}(z,T)$ is computed using cosine similarity.

The Tip-Adapter extends this by constructing a few-shot cache (visualized in supplementary) from the supports $\{(x_j, y_j)\}_{j=1}^N$ using keys $\mathcal{K}\in \mathbb{R}^{N\times D}$ where $\mathcal{K}_j = f_v(x_j)$, and values $\mathcal{V}\in\{0,1\}^{N \times C}$ as one-hot labels. A lightweight adapter $\mathcal{A}:\mathbb{R}^D\rightarrow\mathbb{R}^D$ maps the query feature; cache affinities use cosine similarity $s_j=\text{cos}(\mathcal{A}(z), \mathcal{K}_j)$. Following \cite{tipadapter}, cache logits are computed as $L_{\text{Cache}}(x) = \text{exp}(-\beta(1-s))^T\mathcal{V}$, where $s=[s_1,\dots,s_N]^T$ with $\beta$ is a temperature-like hyperparameter that controls the sharpness of the cosine similarity. At inference, the final logits are: $L_{\text{test}}(x) = L_{\text{ZS}}(x) + \alpha \cdot L_{\text{Cache}}(x)$, where $\alpha\in \mathbb{R}^+$ controls the relative importance of the cache prediction versus the CLIP prior. The only learnable parameter is $\mathcal{A}$. Our goal is to improve the representational power of $\mathcal{A}$ leaving this test-time path unchanged.
\subsection{Train-Time Asymmetric supervision}
\label{sec:framework}
During training, we use a teacher that performs structured, cross-modal reasoning to supervise the student adapter $\mathcal{A}$:
\begin{equation}
\label{eq:train_logits}
L_{\text{train}}(x) = L_{\text{ZS}}(x) + \alpha \cdot L_{\text{Cache}}(x) + \delta \cdot L_{\text{Graph}}(x)
\end{equation}
where $L_{\text{Graph}}$ are teacher logits (Sect.~\ref{sec:teacher}) and $\delta$ balance the supervision. The mixture in \eqref{eq:train_logits} is used only for the training classification loss; gradients update both the student 
and the teacher, refining the latter as an expert during training. At test time, we use the above $L_{\text{test}}$ and discard the teacher.

\begin{figure*}[t!]
\centering
\includegraphics[width=\linewidth]{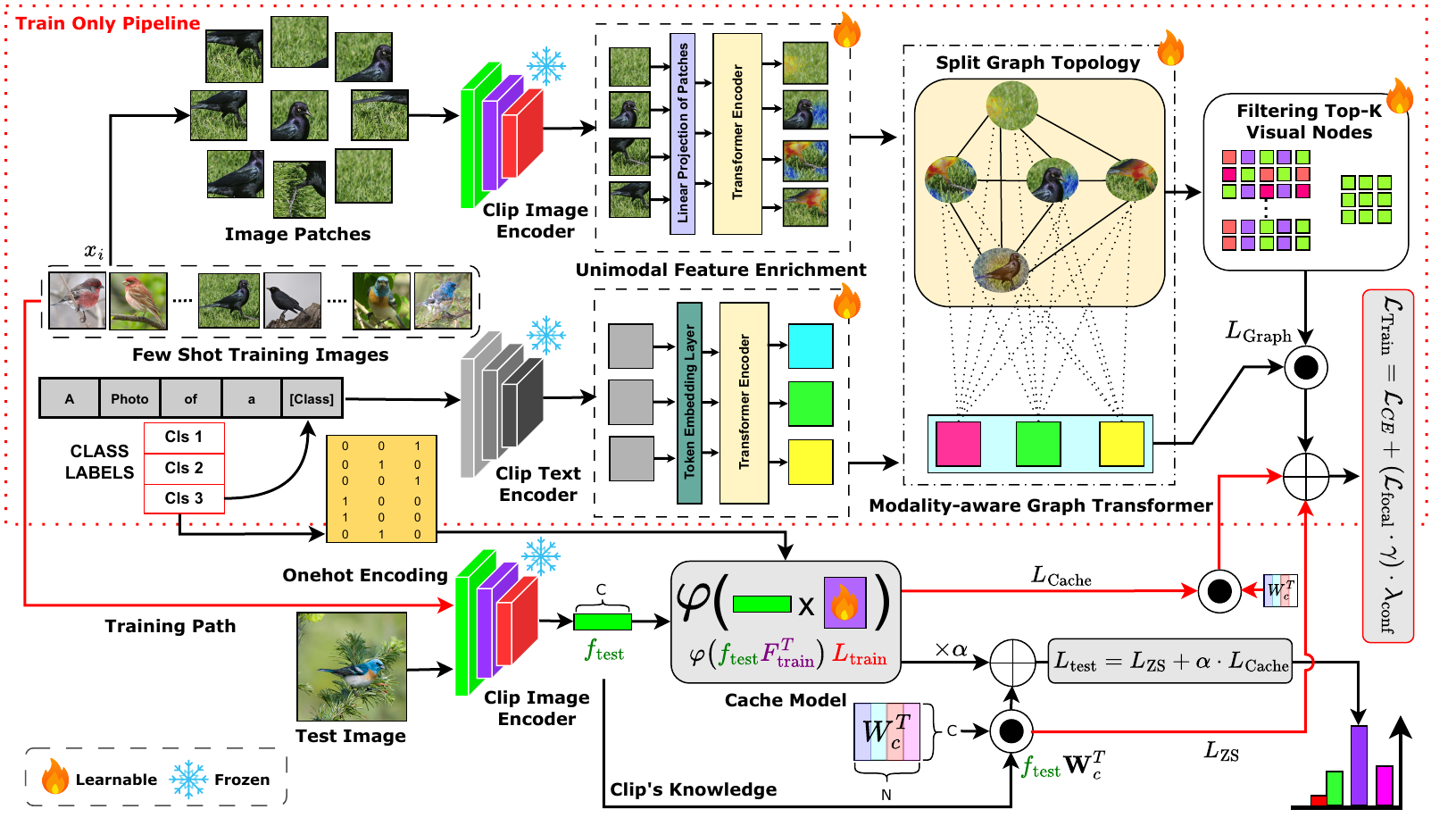}
\caption{Our Training-Only Graph Adapter (\textbf{TOGA}) is an asymmetric supervision pipeline. At \textbf{train time} (red dotted region and \textcolor{red}{$\rightarrow$} denote training only), we use a three-branch ensemble: (1) a frozen $L_\text{ZS}$ branch, (2) a lightweight student (Cache Model) $L_\text{Cache}$, and (3) our powerful, auxiliary graph teacher $L_\text{Graph}$. The teacher enriches multi-scale patches and text embeddings via unimodal Transformers, then constructs a modality-aware graph transformer to perform cross-modal reasoning. A dual-loss objective regularizes the teacher's knowledge into the student's adapter $\mathcal{A}$. At \textbf{test time}, the entire teacher branch is discarded, resulting in zero additional inference cost.}
\label{fig:architecture}
\end{figure*}

\subsection{Teacher: Image$–$Patch$–$Text Graph Topology}
\label{sec:teacher}
Our asymmetric supervision framework comes from its novel auxiliary teacher. As shown in Figure \ref{fig:architecture}, it is a graph-driven model designed to transfer knowledge into the student's adapter $\mathcal{A}$ by exploring fine-grained visual cues and image-patch–text alignments that are invisible to global vectors $f_v$. Unlike the student, which operates on a single global feature $f_v$, the teacher processes a rich, multi-scale patch representation of the image.

\noindent\textbf{Multi-Scale Patch Extraction:} 
To construct the teacher's input, we generate a set of $M$ multi-scale patches from the input image $x$. This set is the union of five distinct views: (1) global view (entire image), (2) local grid $3 \times 3$ (9 patches), (3) medium grid $2 \times 2$ (4 patches), (4) vertical halves (2 patches) and (5) horizontal halves (2 patches). We resize all crops to CLIP’s resolution ($224\times 224$) and pass them through the frozen CLIP image encoder to obtain normalized features $V_{\text{vis}}^{(0)}=\{f_p\}_{p=1}^M$, $f_p\in \mathbb{R}^D$, $||f_p||_2=1$. Similarly, we use the frozen text encoder to obtain normalized class prompts $V_{\text{text}}^{(0)}=\{t_c\}_{c=1}^C$, $t_c\in \mathbb{R}^D$, $||t_c||_2=1$. 

\noindent\textbf{Modality-Specific Projections and Tokens:} 
Before cross-modal reasoning, we learn lightweight projections with a Transformer \cite{vaswani2017attention} (Fig.~\ref{fig:architecture}) to adapt frozen CLIP spaces to a task-specific latent $d_h$. This lets our graph nodes $V_{\text{vis}}^{(0)}$ and $V_{\text{text}}^{(0)}$ first gather within-modality context (e.g., patch$\leftrightarrow$patch co-occurrence; prompt$\leftrightarrow$prompt semantics) before any image-patch-text interaction. 
Given node features $H^{(l)}\in \mathbb{R}^{n\times d_h}$ at layer $l=1,\cdots, L$, multi-head self-attention $\mathcal{H}$ heads compute query $Q_j^{(l)}=H^{(l)}W_Q^j$, key $K_j^{(l)}=H^{(l)}W_K^j$, and value $V_j^{(l)}=H^{(l)}W_V^j$, with $W_Q^j$, $W_K^j$, $W_V^j\in \mathbb{R}^{d_h\times d_k}$, $d_k=d_h/\mathcal{H}$. The attention and head outputs are $\tilde{A}_j^{(l)} = \text{softmax}( Q_j^{(l)}(K_j^{(l)})^T / \sqrt{d_k} )$, $Z_j^{(l)} = \tilde{A}_j^{(l)}V_j^{(l)}$. The heads are concatenated $Z^{(l)} = \text{Concat}(Z_1^{(l)}, ..., Z_{\mathcal{H}}^{(l)})$, then passed through residual, layer normalization (LN), a feed-forward network (FFN):
\begin{equation}
\label{eq:transformer_block}
\begin{split}
H'^{(l)} &= \text{LN}(H^{(l)} + Z^{(l)}) \\
H^{(l+1)} &= \text{LN}(H'^{(l)} + \text{FFN}(H'^{(l)}))
\end{split}
\end{equation}
Applying this stack independently to both modalities ($V_{\text{vis}}^{(0)}$ and $V_{\text{text}}^{(0)}$) yields a contextualized unimodal output that preserves CLIP priors while exposing the local structure. The final outputs $V_{\text{vis}}^{(L)}=H_{\text{vis}}^{(L)}=\{h^{vis}_p\}_{p=1}^M$ and $V_{\text{text}}^{(L)}=H_{\text{text}}^{(L)}=\{h_c^{\text{text}}\}_{c=1}^C$, which serve as inputs to subsequent cross-modal graph reasoning.

\noindent\textbf{Modality-aware Graph Transformer (MGT):} 
Given the enriched unimodal sequences $V_{\text{vis}}^{(L)}$ and $V_{\text{text}}^{(L)}$, we construct our modality-aware graph topology $\mathcal{G}=(\mathcal{N},\mathcal{E})$ with node types $\phi(v)\in\{\text{patch}, \text{text}\}$ and relation (edge) types $\psi(e)\in \mathcal{R}$: $\mathcal{V}=V_{\text{vis}}^{(L)}\bigcup V_{\text{text}}^{(L)}$, $\epsilon=\{(s\xrightarrow{r}t)\}$, $r\in\{\text{\textbf{pp}}, \text{\textbf{pt}}, \text{\textbf{tp}}\}$. Edges \textbf{pp} captures patch$\leftrightarrow$patch, image$\leftrightarrow$patch, and patch$\leftrightarrow$image affinity; \textbf{pt} and \textbf{tp} bind patches/image with textual prompts. $s\in \mathcal{N}$ and $t\in \mathcal{N}$ are the source and target nodes. 

Inspired by inductive bias \cite{hu2020heterogeneous}, we use \textit{type-specific} projections and \textit{relation-specific} parameters to advance the transformer model for our MGT. Let $H_G$ be the number of heads and $d_k^G=d_h/H_G$. At layer $l=1,\cdots, L_G$ for a target node $t$ and its neighbors $s\in \mathcal{N}(t)$ connected via relation $r=\psi(s\rightarrow t)$, head $h\in \{1,\cdots,H_G\}$ uses: (1) Node-type projection: query $Q_t^{(h)}=W_Q^{(h,\phi(t))}h_t^{(l)}$, key $K_s^{(h)}=W_K^{(h,\phi(s))}h_s^{(l)}$, and value $V_s^{(h)}=W_V^{(h,\phi(s))}h_s^{(l)}$. (2) Relational-specific transforms and priors (MGT’s relation-aware keys/values and attention bias): $\tilde{K}_{s\rightarrow t}^{(h)}=W_{K,r}^{(h)}K_s^{(h)}$, $\tilde{V}_{s\rightarrow t}^{(h)}=W_{V,r}^{(h)}V_s^{(h)}$, $b_r^{(h)}\in \mathbb{R}$. Here $W_{(.),r}^{(h)}$ are small relation-type adapters; $b_r^{(h)}$ is an attention bias. (3) Attention with relation bias, relation-gated message aggregation: 

\begin{equation}
\label{eq:hgt_bias}
\begin{aligned}
&e_{s\rightarrow t}^{(h)} = \frac{Q_t^{(h)T} \tilde{K}_{s\rightarrow t}^{(h)}}{\sqrt{d_k^G}} + b_r^{(h)}, \alpha_{s\rightarrow t}^{(h)}=\text{softmax}_{s\in \mathcal{N}(t)}(e_{s\rightarrow t}^{(h)})\\
&m_t^{(h)} = \sum_{s\in\mathcal{N}(t)}\mu_r^{(h)}\alpha_{s\rightarrow t}^{(h)}\tilde{V}_{s\rightarrow t}^{(h)},m_t=\text{Concat}\left(\sum_r\sum_{h=1}^{H_G}m_t^{(h)}\right)\\
&h_t^{(l+1)}=\text{LN}(h_t^{(l)}+FFN^{(\phi(t))}(m_t W_O^{(\phi(t))})
\end{aligned}
\end{equation}
where $\mu_r^{(h)}\in\mathbb{R}_+$ is a learned \textit{relation-level scaling coefficient} that captures the relative importance of each type of relation 
$r$ (e.g., patch-patch vs. patch-text) for head $h$. the type-specific $W_{\{Q,K,V\}}^{(h,\phi)}$, $W_O^{(\phi)}$, and FFN$^{(\phi)}$ preserve modality characteristics (visual patches vs. text prompts), while the relation-specific $W_{(.),r}^{(h)}$, $b_r^{(h)}$, and $\mu_r^{(h)}$ let the transformer weight \textbf{pp} vs \textbf{pt}/\textbf{tp} interactions differently. After layers $L_G$, we obtained deeply contextualized node embeddings $V'_{\text{vis}}=\{h_i^{(L_G)}:\phi(i)=\text{patch/image}\}$, $V'_{\text{text}}=\{h_c^{(L_G)}:\phi(c)=\text{text}\}$. These $V'_{\text{vis}}$ and  $V'_{\text{text}}$ feed our discriminative readout and asymmetric supervision objectives.

\noindent\textbf{Discriminative Node Filtering:} 
We need to combine the final refined visual nodes (patches) $V'_{\text{vis}}=\{h_i\}_{i=1}^P$ from the MGT to represent a final embedding for the given image $x$. Global pooling treats all patches equally and can dilute fine-grained evidence. We therefore perform selective pooling to keep only the most discriminative patches. We learn to filter for the most discriminating nodes using a learnable projection $p\in \mathbb{R}^{d_h}$ and score each node/patch as $s_i = \langle h_i \cdot p\rangle / {\|p\|_2 \|h_i\|_2}$, We rank these nodes using $s_i$ and select the top $\mathbb{N}$ nodes/patches: $V_{\text{vis}}^\mathbb{N}=\text{TopN}\left(\{s_i\}_{i=1}^P, \mathbb{N}\right)$. The features in the selected nodes are aggregated to produce the final teacher's visual feature $f_{\text{graph}}$. The teacher’s logits are then computed in a CLIP-consistent way: $L_{\text{Graph}}(x)_c=\text{cos}\left(f_{\text{graph}}, \hat{t}_c\right)$, $c=1,\cdots, C$ and $V'_{\text{text}}=\{\hat{t}_c\}_{c=1}^C$.

\subsection{Dual-Objective Co-Training}
\label{sec:loss}
We ensure the training-only teacher remains a strong expert while its relational signal is transferred online to the student adapter $\mathcal{A}$ without collapsing into the strong ZS branch. The different logits (ZS, Cache, and Graph) and their combination into final logits $L_{\text{train}}$ are given in Eqn (\ref{eq:train_logits}) in Sec. \ref{sec:framework}. We compute the total loss $\mathcal{L}_{\text{Total}}$ by combining a primary asymmetric supervision term (on the mixture) with a teacher-forcing term (on the teacher alone):
\begin{equation}
\label{eq:total_loss}
\mathcal{L}{_\text{Total}} = \underbrace{\mathcal{L}{_\text{CE}}(L_{\text{train}}, y)}_{\text{joint ensemble loss}} + \lambda \cdot \underbrace{\mathcal{L}{_\text{Focal}}(L{_\text{Graph}},y)}_{\text{teacher forcing}}
\end{equation}
Here $\lambda\ge 0$ sets the strength of teacher forcing on the student's adapter $\mathcal{A}$. We use the standard cross-entropy to compute $\mathcal{L}{_\text{CE}}$ and the gradients $\nabla \mathcal{L}{_\text{CE}}$ propagate to both the student and the teacher. To prevent the teacher from ``coasting'' on ZS/student predictions, we apply \textit{Focal Loss} \cite{lin2017focal} to the teacher logits alone: $p_t = \text{softmax}(L_{\text{Graph}})_y$, $\mathcal{L}{_\text{Focal}} = - (1 - p_t)^{\gamma} \log(p_t)$ with focusing parameter $\gamma\ge 0$. This down-weights easy examples ($p_t\rightarrow 1$) and forces the teacher to allocate capacity to hard, fine-grained cases, improving the quality and stability of the supervisory signal.

The testing pipeline matches the Tip-Adapter baseline (Sec.~\ref{sec:baseline}). Each test image is encoded by CLIP into a single global feature, and the final prediction is computed as $L_{\text{test}} = L_{\text{ZS}} + \alpha \cdot L_{\text{Cache}}$ (Eqn.~\ref{eq:train_logits}, Sec.~\ref{sec:framework}).

%% file: sec/4_experiments.tex
\section{Experimental Results and Discussion}
\label{sec:experiments}

\textbf{Datasets and Protocols.} 
We evaluate on 11 standard benchmarks spanning fine-grained, texture, scene, object, satellite and action categories: FGVC-Aircraft \cite{maji2013fineaircraft}, Flowers102 \cite{nilsback2008automatedflowers}, SUN397 \cite{xiao2010sun}, Food101 \cite{bossard2014food}, CalTech101 \cite{fei2004learning}, UCF101 \cite{soomro2012ucf101}, StanfordCars \cite{krause20133d}, DTD \cite{cimpoi2014describing}, Imagenet \cite{deng2009imagenet}, OxfordPets \cite{parkhi2012cats} and EuroSAT \cite{helber2019eurosat}. Consistent with previous works, we evaluate performance under standard k-shot settings \cite{tipadapter}. For each dataset, we sample $K\in \{1,2,4,8,16\}$ labeled instances per class for training and use the official test/val splits for evaluation. The AdamW optimizer is used with an initial learning rate of $1 \times 10^{-3}$, which decreases according to a cosine annealing schedule. Following the protocols in ~\cite{tipadapter, zhou2022learning}, we repeat the training runs independently three times for each shot configuration and report the average classification accuracy. All experiments were executed on an RTX PRO 6000 (98 GB).

\noindent\textbf{Implementation Details.}
We implement in PyTorch. The backbone is CLIP ViT-B/16 \cite{radford2021clip} with frozen weights, results with other backbones are in supplementary. Data preprocessing follows~\cite{gao2024clipadapter, tipadapter} standard augmentations. Our model depends on four experimentally optimized hyperparameters ($\alpha, \beta, \gamma, \mathbb{N}$); sensitivity and computational complexity analyses are detailed in the supplementary material.

\begin{figure*}[t!]
  \centering
  
  \begin{minipage}[c]{0.75\textwidth} 
    \centering
    \captionof{table}{Comparison of few-shot classification accuracy (\%) on 11 benchmark datasets. We evaluate our method against several SOTA adapter-based approaches. The best performance in each shot-group is marked in \textbf{bold}. Our results are highlighted in blue. \textbf{Dataset abbreviations:} INet (ImageNet), SUN (SUN397), Air (FGVC-Aircraft), Euro (EuroSAT), Cars (Stanford Cars), Food (Food101), Pets (OxfordPets), Flow (Flowers102), Cal (Caltech101), DTD (Describable Textures), UCF (UCF101).}
    \label{tab:comparison}
    
    \resizebox{\linewidth}{!}{%
    \setlength{\tabcolsep}{4pt}
    \begin{tabular}{@{}l l l r r r r r r r r r r r r@{}}
    \toprule
    Shots & Method & Venue & INet & SUN & Air & Euro & Cars & Food & Pets & Flow & Cal & DTD & UCF & Avg \\
    \midrule
    0 & CLIP \textsuperscript{\citep{radford2021clip}} & ICML'22 & 60.3 & 58.5 & 17.1 & 37.5 & 55.7 & 77.3 & 89.1 & 66.0 & 85.9 & 42.2 & 61.5 & 59.2 \\
    \midrule
    \multirow{6}{*}{1}
    & TIP-Adapter-F \textsuperscript{\citep{tipadapter}} & ECCV'22 & 61.1 & 62.5 & 20.2 & 59.5 & 58.9 & 77.5 & 87.0 & 80.0 & 89.3 & 49.6 & 64.9 & 64.6 \\
    & TaskRes \textsuperscript{\citep{yu2023task}} & CVPR'23 & 61.9 & 62.3 & 21.4 & 61.7 & 59.1 & 74.0 & 83.6 & 79.2 & 88.8 & 50.2 & 64.8 & 64.3 \\
    & GraphAdapter\textsuperscript{~\citep{li2023graphadapter}} & NeurIPs'23 & 61.5 & 61.9 & 20.9 & 63.3 & 59.7 & 75.4 & 84.4 & 80.0 & 88.9 & 51.8 & 64.9 & 64.8 \\
    & CLIP-Adapter \textsuperscript{\citep{gao2024clipadapter}} & IJCV'24 & 61.2 & 61.3 & 17.5 & 61.4 & 55.1 & 76.8 & 86.0 & 73.5 & 88.6 & 45.8 & 62.2 & 62.7 \\
    & CLAP\textsuperscript{\citep{silva2024closer}} & CVPR'24 & 58.5 & 61.1 & 20.6 & 59.2 & 56.3 & 73.0 & 83.6 & 79.9 & 88.4 & 47.5 & 62.5 & 62.8 \\
    & CCA\textsuperscript{\citep{jiang2025causal}} & ICCV'25 & 61.5 & 63.8 & 22.5 & 67.0 & 60.0 & 77.8 & 86.9 & 81.0 & 89.9 & 51.0 & 66.3 & 66.3 \\
    \cmidrule(lr){2-15}
    & \textbf{TOGA (Ours)} & - & \cellcolor{blue!15} \textbf{69.2} & \cellcolor{blue!15}\textbf{68.1} & \cellcolor{blue!15}\textbf{31.0} & \cellcolor{blue!15}\textbf{67.4} & \cellcolor{blue!15}\textbf{69.1} & \cellcolor{blue!15}\textbf{86.2} & \cellcolor{blue!15}\textbf{91.2} & \cellcolor{blue!15}\textbf{88.2} & \cellcolor{blue!15}\textbf{94.3} & \cellcolor{blue!15}\textbf{55.2} & \cellcolor{blue!15}\textbf{74.5} & \cellcolor{blue!15}\textbf{72.2} \\
    \midrule
    \multirow{6}{*}{2}
    & TIP-Adapter-F \textsuperscript{\citep{tipadapter}} & ECCV'22 & 61.7 & 63.6 & 23.2 & 66.1 & 61.5 & 77.8 & 87.0 & 82.3 & 89.7 & 53.7 & 66.4 & 66.6 \\
    & TaskRes \textsuperscript{\citep{yu2023task}} & CVPR'23 & 61.9 & 64.9 & 24.1 & 65.8 & 63.7 & 75.2 & 84.6 & 86.6 & 90.3 & 55.1 & 70.0 & 67.5 \\
    & GraphAdapter\textsuperscript{~\citep{li2023graphadapter}} & NeurIPs'23 & 62.3 & 64.6 & 23.8 & 67.3 & 63.2 & 76.3 & 86.3 & 85.6 & 90.2 & 55.7 & 69.5 & 67.7 \\
    & CLIP-Adapter \textsuperscript{\citep{gao2024clipadapter}} & IJCV'24 & 61.5 & 63.3 & 20.1 & 63.9 & 58.7 & 77.2 & 86.7 & 81.6 & 89.4 & 51.5 & 67.1 & 65.5 \\
    & CLAP\textsuperscript{\citep{silva2024closer}} & CVPR'24 & 58.5 & 63.3 & 23.2 & 65.6 & 61.4 & 74.9 & 84.9 & 84.2 & 89.8 & 53.0 & 67.8 & 66.1 \\
    & CCA\textsuperscript{\citep{jiang2025causal}} & ICCV'25 & 62.1 & 66.3 & 25.0 & 70.0 & 64.0 & 77.9 & 87.9 & 88.0 & 91.0 & 55.0 & 69.3 & 68.9 \\
    \cmidrule(lr){2-15}
    & \textbf{TOGA (Ours)} & - & \cellcolor{blue!15}\textbf{69.9} & \cellcolor{blue!15}\textbf{70.7} & \cellcolor{blue!15}\textbf{34.8} & \cellcolor{blue!15}\textbf{74.9} & \cellcolor{blue!15}\textbf{72.8} & \cellcolor{blue!15}\textbf{86.5} & \cellcolor{blue!15}\textbf{92.2} & \cellcolor{blue!15}\textbf{91.9} & \cellcolor{blue!15}\textbf{95.0} & \cellcolor{blue!15}\textbf{58.4} & \cellcolor{blue!15}\textbf{77.9} & \cellcolor{blue!15}\textbf{75.0} \\
    \midrule
    \multirow{6}{*}{4}
    & TIP-Adapter-F \textsuperscript{\citep{tipadapter}} & ECCV'22 & 62.5 & 66.2 & 25.8 & 74.1 & 64.6 & 78.2 & 87.5 & 88.8 & 90.6 & 57.4 & 70.6 & 69.7 \\
    & TaskRes \textsuperscript{\citep{yu2023task}} & CVPR'23 & 63.6 & 67.3 & 25.7 & 73.8 & 67.4 & 76.1 & 86.3 & 90.2 & 91.0 & 60.7 & 70.9 & 70.3 \\
    & GraphAdapter\textsuperscript{~\citep{li2023graphadapter}} & NeurIPs'23 & 63.1 & 66.7 & 27.0 & 75.2 & 66.5 & 76.8 & 86.6 & 89.9 & 91.0 & 59.6 & 71.5 & 70.3 \\
    & CLIP-Adapter \textsuperscript{\citep{gao2024clipadapter}} & IJCV'24 & 62.7 & 66.0 & 22.6 & 73.4 & 62.4 & 77.9 & 87.5 & 87.2 & 90.0 & 56.9 & 69.1 & 68.7 \\
    & CLAP \textsuperscript{\citep{silva2024closer}} & CVPR'24 & 60.7 & 65.9 & 25.6 & 73.1 & 65.5 & 75.9 & 86.5 & 87.6 & 90.6 & 58.8 & 69.8 & 69.1 \\
    & CCA\textsuperscript{\citep{jiang2025causal}} & ICCV'25 & 63.3 & 69.0 & 29.0 & 80.1 & 68.0 & 78.2 & 88.1 & 91.1 & 92.0 & 63.0 & 72.1 & 72.2 \\
    \cmidrule(lr){2-15}
    & \textbf{TOGA (Ours)} & - & \cellcolor{blue!15}\textbf{71.0} & \cellcolor{blue!15}\textbf{73.2} & \cellcolor{blue!15}\textbf{38.3} & \cellcolor{blue!15}\textbf{80.3} & \cellcolor{blue!15}\textbf{76.7} & \cellcolor{blue!15}\textbf{86.7} & \cellcolor{blue!15}\textbf{92.7} & \cellcolor{blue!15}\textbf{96.4} & \cellcolor{blue!15}\textbf{95.7} & \cellcolor{blue!15}\textbf{64.5} & \cellcolor{blue!15}\textbf{81.7} & \cellcolor{blue!15}\textbf{77.9} \\
    \midrule
    \multirow{6}{*}{8}
    & TIP-Adapter-F \textsuperscript{\citep{tipadapter}} & ECCV'22 & 64.0 & 68.9 & 30.2 & 77.9 & 69.2 & 78.6 & 88.1 & 91.5 & 91.4 & 62.7 & 74.2 & 72.4 \\
    & TaskRes \textsuperscript{\citep{yu2023task}} & CVPR'23 & 64.7 & 68.7 & 31.5 & 79.3 & 71.8 & 76.4 & 87.2 & 94.7 & 92.4 & 64.8 & 75.3 & 73.3 \\
    & GraphAdapter\textsuperscript{~\citep{li2023graphadapter}} & NeurIPs'23 & 64.2 & 68.9 & 31.4 & 80.2 & 70.5 & 77.7 & 87.6 & 94.1 & 92.4 & 64.5 & 75.7 & 73.4 \\
    & CLIP-Adapter \textsuperscript{\citep{gao2024clipadapter}} & IJCV'24 & 62.7 & 67.5 & 26.2 & 77.9 & 67.9 & 78.0 & 87.6 & 91.7 & 91.4 & 61.0 & 73.3 & 71.4 \\
    & CLAP\textsuperscript{\citep{silva2024closer}} & CVPR'24 & 62.9 & 68.6 & 28.9 & 76.7 & 70.3 & 77.4 & 87.7 & 92.1 & 91.4 & 63.2 & 73.3 & 72.1 \\
    & CCA\textsuperscript{\citep{jiang2025causal}} & ICCV'25 & 64.9 & 71.0 & 35.0 & 83.1 & 74.3 & 79.2 & 89.0 & 94.4 & 92.5 & 65.0 & 77.5 & 75.0 \\
    \cmidrule(lr){2-15}
    & \textbf{TOGA (Ours)} & - & \cellcolor{blue!15} \textbf{71.5} & \cellcolor{blue!15}\textbf{75.1} & \cellcolor{blue!15}\textbf{44.2} & \cellcolor{blue!15}\textbf{84.1} & \cellcolor{blue!15}\textbf{78.2} & \cellcolor{blue!15}\textbf{86.8} & \cellcolor{blue!15}\textbf{93.4} & \cellcolor{blue!15}\textbf{97.3} & \cellcolor{blue!15}\textbf{95.8} & \cellcolor{blue!15}\textbf{69.6} & \cellcolor{blue!15}\textbf{83.9} & \cellcolor{blue!15}\textbf{80.0} \\
    \midrule
    \multirow{7}{*}{16}
    & TIP-Adapter-F \textsuperscript{\citep{tipadapter}} & ECCV'22 & 65.5 & 71.5 & 35.6 & 84.5 & 75.7 & 79.4 & 89.7 & 94.8 & 92.9 & 65.6 & 78.0 & 75.7 \\
    & TaskRes \textsuperscript{\citep{yu2023task}} & CVPR'23 & 63.7 & 70.7 & 36.3 & 84.0 & 76.8 & 77.6 & 87.8 & 96.0 & 93.4 & 67.1 & 78.0 & 75.8 \\
    & GraphAdapter\textsuperscript{~\citep{li2023graphadapter}} & NeurIPs'23 & 65.7 & 71.2 & 36.9 & 85.3 & 76.2 & 78.6 & 88.6 & 96.2 & 93.3 & 67.6 & 78.8 & 76.2 \\
    & CLIP-Adapter \textsuperscript{\citep{gao2024clipadapter}} & IJCV'24 & 63.6 & 69.6 & 32.1 & 84.4 & 74.0 & 78.2 & 87.8 & 93.9 & 92.5 & 66.0 & 76.8 & 74.4 \\
    & CLAP \textsuperscript{\citep{silva2024closer}} & CVPR'24 & 65.0 & 70.8 & 33.6 & 80.1 & 75.1 & 78.5 & 88.5 & 94.2 & 91.9 & 66.4 & 76.3 & 74.6 \\
    & CCA\textsuperscript{\citep{jiang2025causal}} & ICCV'25 & 66.0 & 72.2 & 42.0 & 85.3 & 79.0 & 79.8 & 90.9 & 95.2 & 93.5 & 70.0 & 80.0 & 77.6 \\
    \cmidrule(lr){2-15}
    & \textbf{TOGA (Ours)} & - & \cellcolor{blue!15}\textbf{72.3} & \cellcolor{blue!15}\textbf{76.2} & \cellcolor{blue!15} \textbf{48.4} & \cellcolor{blue!15}\textbf{89.4} & \cellcolor{blue!15} \textbf{85.3} & \cellcolor{blue!15} \textbf{87.5} & \cellcolor{blue!15} \textbf{93.4} & \cellcolor{blue!15} \textbf{98.3} & \cellcolor{blue!15}\textbf{96.3} & \cellcolor{blue!15} \textbf{73.6} & \cellcolor{blue!15}\textbf{84.9} & \cellcolor{blue!15}\textbf{82.3} \\
    \bottomrule
    \end{tabular}%
    } 
  \end{minipage}
  \hfill
  \begin{minipage}[c]{0.23\textwidth} 
    \centering
    
    \includegraphics[width=\textwidth]{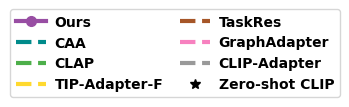}
    
    \begin{subfigure}[b]{\textwidth}
      \includegraphics[width=\textwidth]{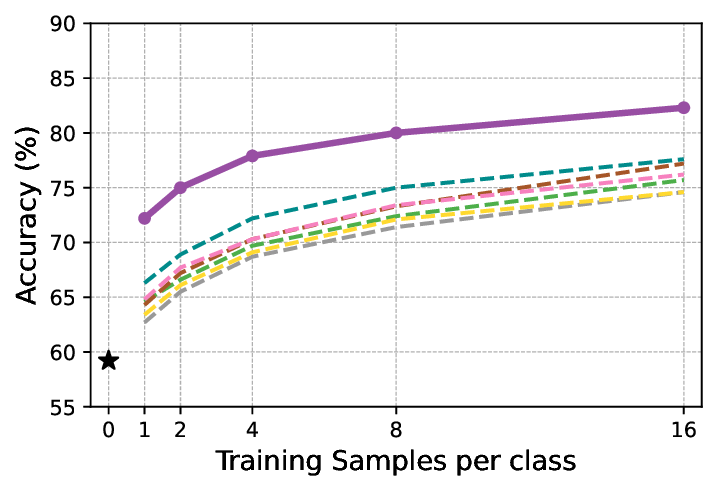}
      \caption{Average}
      \label{fig:average}
    \end{subfigure}
    
    
    \begin{subfigure}[b]{\textwidth}
      \includegraphics[width=\textwidth]{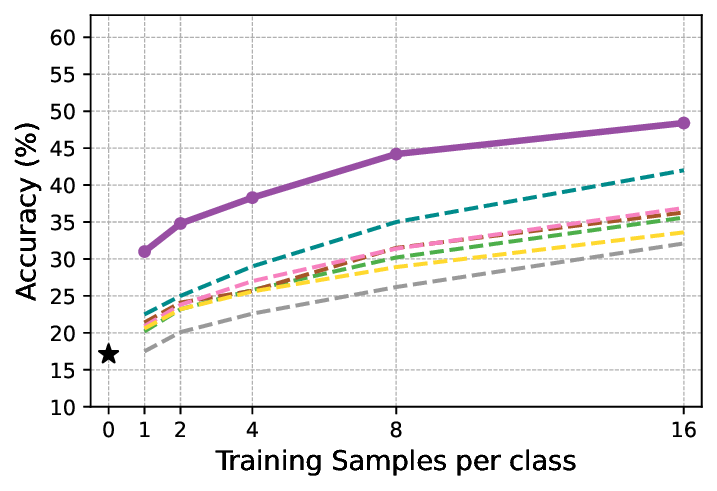}
      \caption{Aircraft}
      \label{fig:caltech101}
    \end{subfigure}
    
    
    \begin{subfigure}[b]{\textwidth}
      \includegraphics[width=\textwidth]{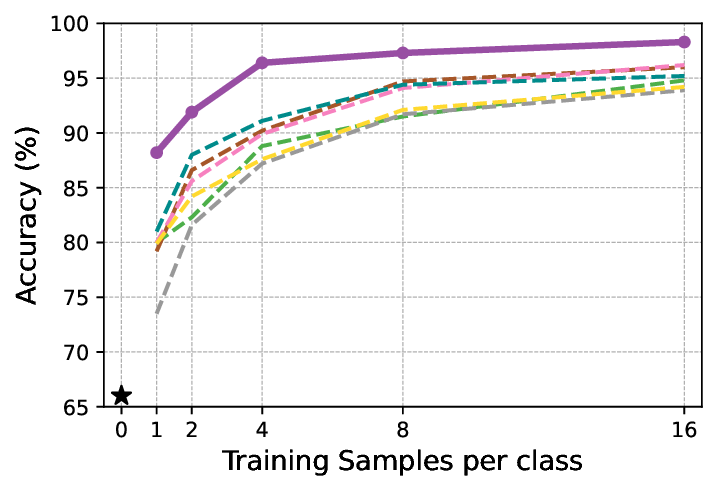}
      \caption{Flowers102}
      \label{fig:Flowers102}
    \end{subfigure} 
    
    
    \begin{subfigure}[b]{\textwidth}
      \includegraphics[width=\textwidth]{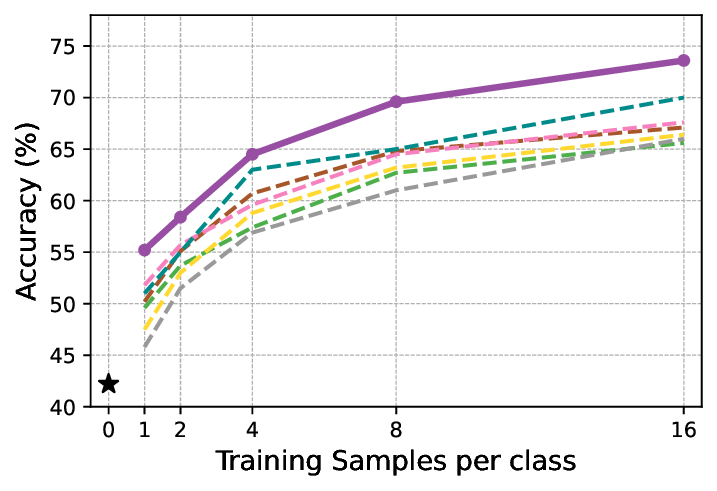}
      \caption{DTD}
      \label{fig:DTD}
    \end{subfigure}
    
    \captionof{figure}{Comparison of few-shot accuracy (\%) on three benchmarks and the 11-dataset average. More results in the supplementary.}
    \label{fig:fewshot}
  \end{minipage}
  
\end{figure*}

\noindent\textbf{Comparison with State-of-the-Art (SOTA).} Table~\ref{tab:comparison} compares our method to leading adapter families, including global-feature (TIP-Adapter-F~\citep{tipadapter}, TaskRes~\citep{yu2023task}), patch-based (GraphAdapter~\citep{li2023graphadapter}), and prompt learning methods (compared in supplementary). Our method establishes a new SOTA across all 11 benchmarks and shot settings while preserving the exact inference path, latency, and parameter count of Tip-Adapter-F. In contrast, patch-level methods like GraphAdapter require permanent test-time computation; our training-only graph teacher avoids this trade-off.



  

\noindent\textbf{Extreme Low-Shot Regime (1-4 Shots).} Strongest gains when data are scarcest. The largest separations emerge in the low-data regime, where global adapters struggle to form stable prototypes. Relative to the nearest strong competitor (a cache-centric adapter we denote CCA), our method improves the 1-shot average by +5.9\% (72.2\% vs 66.3\%), 2-shot by +6.1\% (75.0\% vs 68.9\%), and 4-shot by +5.7\% (77.9\% vs 72.2\%).
Fine-grained categories amplify the effect: on Aircraft, we improve over CCA by +8.5\% (1-shot), +9.8\% (2-shot; 34.8\% vs 25.0\%), and +9.3\% (4-shot; 38.3\% vs 29.0\%). This consistent pattern indicates that relying on a single global vector makes prototypes highly sensitive to sample bias in few-shot settings.  Our image-patch–text graph provides relational supervision (Sect. 3.3), yielding prototypes that better capture class-defining local evidence, even from one labeled example.

\noindent\textbf{Generalization and Scalability.} Datasets like \textit{Stanford Cars} and \textit{Flowers102} demand attention to parts and textures that global pooling dilutes. Through asymmetric supervision by a patch-aware, cross-modal teacher, we close this gap. For example, on \textit{Flowers102} (16-shot) we reach 98.3\%, approaching saturation. Furthermore, on \textit{EuroSAT} (satellite imagery), our method scales to \textbf{89.4\%} (16-shot), while TIP-Adapter-F plateaus at 84.5\% , GraphAdapter/CCA plateaus at 85.3\%. This indicates that our bi-modal graph reasoning provides a more robust feature alignment when the downstream task diverges significantly from CLIP's pre-training distribution. Finally, comparing directly with GraphAdapter~\citep{li2023graphadapter}, our method consistently achieves higher accuracy (e.g., 82.3\% vs. 76.2\% average at 16-shot) without requiring explicit graph computation at test time. This confirms that our asymmetric supervision strategy effectively ``compiles'' the relational knowledge of modality-aware graph into the lightweight student adapter, achieving superior performance that is fundamentally more efficient at inference.

\begin{table}[t]
    \scriptsize
    \centering
    \caption{Out-of-distribution (OOD) robustness accuracy (\%) on ImageNet and its shifted variants; Avg: mean across five datasets.}
    \label{tab:ood_results}
    \resizebox{\linewidth}{!}{
    
    \begin{tabular}{l c ccccc}
        \toprule
        \textbf{Method} & \textbf{ImageNet} & \textbf{-R} & \textbf{-A} & \textbf{-V2} & \textbf{-Sketch} & \textbf{Avg} \\
        \midrule
        CLIP \cite{radford2021clip} & 66.7 & 74.0 & 47.8 & 60.8 & 46.2 & 59.1 \\
        CoOp \cite{zhou2022learning} & 71.5 & 75.2 & 49.7 & 64.2 & 48.0 & 61.1 \\
        CoCoOp \cite{zhou2022conditional} & 71.0 & 76.2 & 50.6 & 64.1 & 48.8 & 62.1 \\
        MaPLe \cite{khattak2023maple} & 70.7 & 77.0 & \textbf{50.9} & 64.1 & 49.2 & 62.4 \\
        ProGrad \cite{zhu2023prompt} & 72.2 & 74.6 & 49.4 & 64.7 & 47.6 & 61.7 \\
        KgCoOp \cite{yao2023visual} & 71.2 & 76.7 & 50.7 & 64.1 & 49.0 & 62.3 \\
        TransCLIP \cite{zanella2024boosting} & 70.3 & 75.0 & 49.5 & 62.3 & 49.7 & 61.4 \\
        \midrule
        \rowcolor{blue!15} 
        \textbf{TOGA (Ours)} & \textbf{72.3} & \textbf{77.4} & 49.8 & \textbf{65.9} & \textbf{49.9} & \textbf{63.1} \\
        \bottomrule
    \end{tabular}
    }
\end{table}

\noindent\textbf{OOD Generalization Analysis.} 
We further evaluate the robustness on ImageNet and four standard distribution-shift benchmarks: ImageNet-R, ImageNet-A, ImageNet-V2, and ImageNet-Sketch. Table \ref{tab:ood_results} shows that our method achieves the \textit{best overall average} (\textbf{63.1}), improving zero-shot CLIP by +4.0\% (59.1$\rightarrow$63.1) and outperforms the strong prompt-learning baselines. In particular, we obtain the best results on \textbf{IN (72.3)}, \textbf{R (77.4)}, \textbf{V2 (65.9)}, and \textbf{Sk (49.9)}, while remaining competitive on \textbf{A (49.8)}. These results indicate that our few-shot adaptation does not overfit the support set or erode CLIP’s robustness under shift.

We attribute this to the fact that the CLIP backbone is frozen and inference retains the zero-shot prior, $L_{\text{test}}=L_{\text{ZS}}+\alpha L_{\text{Cache}}$. The graph teacher is used only during training to distill patch-level and cross-modal relational knowledge into the cache adapter, and is fully discarded at test time. Thus, our method improves the adaptation performance while preserving a strong OOD generalization.
\subsection{Ablation Study}

\begin{table}[t]
\centering
\caption{Ablation of the dual-objective co-training strategy. $\mathcal{L}_{CE}$ denotes the cross-entropy loss on the final logits. $\mathcal{L}_{CE}^{Graph}$ and $\mathcal{L}_{Focal}^{Graph}$ are the auxiliary losses applied to the teacher.
Dataset abbreviation follows the same as table \ref{tab:comparison}.}
\label{tab:loss_ablation}
\scriptsize 
\resizebox{\linewidth}{!}{
\begin{tabular}{@{}l l c c c c c@{}}
\toprule
& Loss Objective & 1 Shot & 2 Shot & 4 Shot & 8 Shot & 16 Shot \\
\midrule
\multirow{3}{*}{\rotatebox[origin=c]{90}{\textit{Euro}}}
& $\mathcal{L}_{CE}$ & 65.1 & 73.1 & 77.9 & 83.2 & 88.1 \\
& $\mathcal{L}_{CE} + \mathcal{L}_{CE}^{Graph}$ & 67.1 & 74.5 & 79.3 & 82.2 & 87.5 \\
& \cellcolor{gray!15} $\mathcal{L}_{CE} + \mathcal{L}_{Focal}^{Graph}$ & \cellcolor{gray!15}\textbf{67.4} & \cellcolor{gray!15}\textbf{74.9} & \cellcolor{gray!15}\textbf{80.3} & \cellcolor{gray!15}\textbf{84.1} & \cellcolor{gray!15}\textbf{89.4} \\
\midrule
\multirow{3}{*}{\rotatebox[origin=c]{90}{\textit{Cal}}}
& $\mathcal{L}_{CE}$ & 94.0 & 94.3 & 95.4 & 95.5 & 95.8 \\
& $\mathcal{L}_{CE} + \mathcal{L}_{CE}^{Graph}$ & 94.0 & 94.6 & 95.4 & 95.5 & 96.1 \\
& \cellcolor{gray!15} $\mathcal{L}_{CE} + \mathcal{L}_{Focal}^{Graph}$ & \cellcolor{gray!15}\textbf{94.3} & \cellcolor{gray!15}\textbf{95.0} & \cellcolor{gray!15}\textbf{95.7} & \cellcolor{gray!15}\textbf{95.8} & \cellcolor{gray!15}\textbf{96.3} \\
\midrule
\multirow{3}{*}{\rotatebox[origin=c]{90}{\textit{Food}}}
& $\mathcal{L}_{CE}$ & 86.0 & 86.2 & 86.5 & 86.6 & 87.3 \\
& $\mathcal{L}_{CE} + \mathcal{L}_{CE}^{Graph}$ & 86.0 & 86.2 & 86.5 & 86.7 & 87.3 \\
& \cellcolor{gray!15} $\mathcal{L}_{CE} + \mathcal{L}_{Focal}^{Graph}$ & \cellcolor{gray!15}\textbf{86.2} & \cellcolor{gray!15}\textbf{86.5} & \cellcolor{gray!15}\textbf{86.7} & \cellcolor{gray!15}\textbf{86.8} & \cellcolor{gray!15}\textbf{87.5} \\
\bottomrule
\end{tabular}
}
\end{table}

\noindent\textbf{Effectiveness of Dual-Objective Co-Training.} Table~\ref{tab:loss_ablation} ablates the teacher's graph-based objective, $\mathcal{L}^{Graph}$. With only $\mathcal{L}_{CE}$, the teacher merely tracks the ZS+student mixture and provides a weak supervisory signal with limited low-shot gains. Adding a standard $\mathcal{L}_{CE}^{Graph}$ strengthens low-shot performance but creates gradient conflicts at higher shots as the teacher overfits easy examples. Our proposed $\mathcal{L}_{Focal}^{Graph}$ mitigates this by down-weighting easy instances, forcing the teacher to focus on hard, fine-grained cases. This produces a more stable, distinctive supervisory signal and the most consistent gains across all shots, especially on domains far from the pre-training distribution.

\begin{table}[ht]
\centering
\caption{Global pooling ($\sum_{\text{All}}$) vs Top-$\mathbb{N}$ filtering ($\sum_{\text{Top-}\mathbb{N}}$). The Top-$\mathbb{N}$ strategy explicitly filters out non-discriminative patches.}
\label{tab:topk_pooling}
\setlength{\tabcolsep}{5pt} 
\scriptsize
\resizebox{\linewidth}{!}{
\begin{tabular}{llccccc}
\toprule
 & Pooling Strategy & 1 Shot & 2 Shot & 4 Shot & 8 Shot & 16 Shot \\ 
\midrule

\multirow{4}{*}{\rotatebox{90}{\textit{EuroSAT}}} 
 & $\sum_{\text{All}}$ (Global) & 63.4 & 72.2 & 79.5 & 83.4 & 88.7 \\
 & $\sum_{\text{Top-}\mathbb{N}= 25}$ & 65.3 & 73.3 & 79.9 & 83.7 & 88.9 \\
 & \cellcolor{gray!15}$\sum_{\text{Top-}\mathbb{N}= 50}$ & \cellcolor{gray!15}\textbf{67.4} & \cellcolor{gray!15}\textbf{74.9} & \cellcolor{gray!15}\textbf{80.3} & \cellcolor{gray!15}\textbf{84.1} & \cellcolor{gray!15}\textbf{89.4} \\
 & $\sum_{\text{Top-}\mathbb{N}= 75}$ & 64.9 & 73.0 & 79.6 & 83.4 & 88.8 \\ 
\midrule

\multirow{4}{*}{\rotatebox{90}{\textit{Caltech}}} 
 & $\sum_{\text{All}}$ (Global) & 94.1 & 94.6 & 95.5 & 95.6 & 96.2 \\
 & $\sum_{\text{Top-}\mathbb{N}= 25}$ & 94.2 & 94.8 & 95.6 & 95.8 & 96.1 \\
 & \cellcolor{gray!15}$\sum_{\text{Top-}\mathbb{N}= 50}$ & \cellcolor{gray!15}\textbf{94.3} & \cellcolor{gray!15}95.0 & \cellcolor{gray!15}\textbf{95.7} & \cellcolor{gray!15}95.8 & \cellcolor{gray!15}\textbf{96.3} \\
 & $\sum_{\text{Top-}\mathbb{N}= 75}$ & 94.3 & \textbf{95.1} & 95.6 & \textbf{95.9} & 96.2 \\ 
\midrule

\multirow{4}{*}{\rotatebox{90}{\textit{Food}}} 
 & $\sum_{\text{All}}$ (Global) & 86.1 & 86.4 & 86.6 & 86.6 & 87.4 \\
 & $\sum_{\text{Top-}\mathbb{N}= 25}$ & 86.1 & 86.5 & 86.7 & 86.8 & 87.4 \\
 & \cellcolor{gray!15}$\sum_{\text{Top-}\mathbb{N}=50}$ & \cellcolor{gray!15}\textbf{86.2} & \cellcolor{gray!15}\textbf{86.5} & \cellcolor{gray!15}\textbf{86.7} & \cellcolor{gray!15}\textbf{86.8} & \cellcolor{gray!15}87.5 \\
 & $\sum_{\text{Top-}\mathbb{N}= 75}$ & 86.2 & 86.4 & 86.7 & 86.8 & \textbf{87.6} \\ 
\bottomrule
\end{tabular}
}
\end{table}

\noindent\textbf{Effectiveness of Discriminative Node Filtering.} We validate our aggregation mechanism in Table~\ref{tab:topk_pooling}, comparing global sum pooling ($\sum_{\text{All}}$) against our Top-$\mathbb{N}$ filtering. The $\sum_{\text{All}}$ baseline is non-discriminative; it aggregates all node features, allowing background noise to dilute the signal from small foreground targets, which is particularly detrimental in low-shot settings on datasets like EuroSAT. Our $\sum_{\text{Top-}\mathbb{N}}$ filtering resolves this by learning to retain discriminative foreground patches while suppressing background noise (see Figure~\ref{fig:topk}). This yields a cleaner, more robust signal for asymmetric supervision. The performance peak at $\mathbb{N}=50\%$ supports this and highlights a key trade-off: limited filtering ($\mathbb{N}=75\%$) retains noise, while excessive filtering ($\mathbb{N}=25\%$) discards relevant features.

\begin{figure}[t!]
\centering\includegraphics[width=1.0\linewidth]{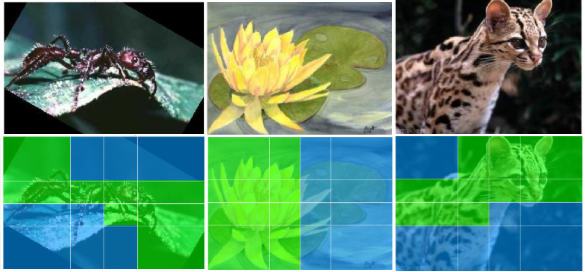} 
\caption{\textbf{Qualitative visualization of Discriminative Node Filtering.} Our Top-$\mathbb{N}$ filtering learns to retain high-scoring, discriminative foreground patches (Green) while suppressing non-informative background nodes (Blue). This resolves feature dilution inherent in global pooling, yielding a cleaner signal by focusing on key object parts (e.g., ant's head, flower pistil, cat's eyes). Samples from Caltech-101; more visualizations in supplementary.}
\label{fig:topk}
\label{fig:topk}
\end{figure}

\begin{table}[t]
\centering
\caption{Ablation of patch grid granularity. Comparison of fixed-scale grids vs MultiScale strategy (18 patches). }
\label{tab:patching_ablation}
\setlength{\tabcolsep}{5pt} 
\scriptsize
\resizebox{\linewidth}{!}{
\begin{tabular}{@{}l l c c c c c@{}}
\toprule
& Patch Strategy & 1 Shot & 2 Shot & 4 Shot & 8 Shot & 16 Shot \\
\midrule
\multirow{4}{*}{\rotatebox[origin=c]{90}{\textit{Euro}}}
& Grid $2\times2$ & 65.0 & 72.9 & 80.0 & 83.7 & 88.8 \\
& Grid $3\times3$ & 61.7 & 72.5 & 78.8 & 83.6 & 88.8 \\
& Grid $4\times4$ & 66.6 & 74.5 & 77.9 & 83.7 & 88.9 \\
& \cellcolor{gray!15}\textbf{MultiScale} & \cellcolor{gray!15}\textbf{67.4} & \cellcolor{gray!15}\textbf{74.9} & \cellcolor{gray!15}\textbf{80.3} & \cellcolor{gray!15}\textbf{84.1} & \cellcolor{gray!15}\textbf{89.4} \\
\midrule
\multirow{4}{*}{\rotatebox[origin=c]{90}{\textit{Cal}}}
& Grid $2\times2$ & 93.7 & 94.5 & 95.5 & 95.6 & 96.2 \\
& Grid $3\times3$ & 93.8 & 94.6 & 95.6 & 95.6 & 96.0 \\
& Grid $4\times4$ & 94.2 & 94.7 & 95.6 & 95.6 & 95.9 \\
& \cellcolor{gray!15}\textbf{MultiScale} & \cellcolor{gray!15}\textbf{94.3} & \cellcolor{gray!15}\textbf{95.0} & \cellcolor{gray!15}\textbf{95.7} & \cellcolor{gray!15}\textbf{95.8} & \cellcolor{gray!15}\textbf{96.3} \\
\midrule
\multirow{4}{*}{\rotatebox[origin=c]{90}{\textit{Food}}}
& Grid $2\times2$ & 86.0 & 86.3 & 86.5 & 86.6 & 87.3 \\
& Grid $3\times3$ & 86.1 & 86.4 & 86.6 & 86.6 & 87.3 \\
& Grid $4\times4$ & 86.0 & 86.4 & 86.6 & 86.6 & 87.4 \\
& \cellcolor{gray!15}\textbf{MultiScale} & \cellcolor{gray!15}\textbf{86.2} & \cellcolor{gray!15}\textbf{86.5} & \cellcolor{gray!15}\textbf{86.7} & \cellcolor{gray!15}\textbf{86.8} & \cellcolor{gray!15}\textbf{87.5} \\
\bottomrule
\end{tabular}
}
\end{table}

\noindent\textbf{Impact of Multi-Scale Patch Hierarchy.} We analyze the teacher's visual granularity in Table~\ref{tab:patching_ablation} (multi-scale patch count ablation in supplementary). Any fixed-scale patch strategy, such as $3 \times 3$ or $4 \times 4$, introduces an inherent trade-off. On datasets with heterogeneous scales like EuroSAT, fine-grained grids (e.g., $4 \times 4$) are effective for local textures but fail to capture larger semantic structures, while coarse grids (e.g., $3 \times 3$) break the context of small-scale objects. Our MultiScale approach is designed to resolve this. By providing the teacher with a heterogeneous set of views, spanning fine-grained patches to coarse and global views, it can reason about local details and global context simultaneously. This yields a more comprehensive feature representation for asymmetric supervision, validated by its superior performance, especially in low-shot scenarios.

\begin{table}[t]
\centering
\caption{Impact of Modality-aware Graph Transformer (MGT).}
\label{tab:mgt_ablation}
\scriptsize
\resizebox{\linewidth}{!}{ 
\begin{tabular}{@{}l l c c c c c@{}}
\toprule
& Configuration & 1 Shot & 2 Shot & 4 Shot & 8 Shot & 16 Shot \\
\midrule
\multirow{2}{*}{\rotatebox[origin=c]{90}{\textit{Euro}}}
& w/o MGT & 61.9 & 71.5 & 77.2 & 80.3 & 85.7 \\
& \cellcolor{gray!15}w/ MGT & \cellcolor{gray!15}\textbf{67.4} & \cellcolor{gray!15}\textbf{74.9} & \cellcolor{gray!15}\textbf{80.3} & \cellcolor{gray!15}\textbf{84.1} & \cellcolor{gray!15}\textbf{89.4} \\
\midrule
\multirow{2}{*}{\rotatebox[origin=c]{90}{\textit{Cal}}}
& w/o MGT & 94.0 & 94.6 & 95.4 & 95.2 & 95.8 \\
& \cellcolor{gray!15}w/ MGT & \cellcolor{gray!15}\textbf{94.3} & \cellcolor{gray!15}\textbf{95.0} & \cellcolor{gray!15}\textbf{95.7} & \cellcolor{gray!15}\textbf{95.8} & \cellcolor{gray!15}\textbf{96.3} \\
\midrule
\multirow{2}{*}{\rotatebox[origin=c]{90}{\textit{Food}}}
& w/o MGT & 86.0 & 86.3 & 86.5 & 86.4 & 87.2 \\
& \cellcolor{gray!15}w/ MGT & \cellcolor{gray!15}\textbf{86.2} & \cellcolor{gray!15}\textbf{86.5} & \cellcolor{gray!15}\textbf{86.7} & \cellcolor{gray!15}\textbf{86.8} & \cellcolor{gray!15}\textbf{87.5} \\
\bottomrule
\end{tabular}
}
\end{table}

\noindent\textbf{Impact of Modality-aware Graph Transformer (MGT).} 
Table~\ref{tab:mgt_ablation} isolates the contribution of the proposed MGT. Removing MGT leaves only unimodal features and consistently degrades performance, especially on the domain-shifted EuroSAT benchmark (e.g., 67.4$\rightarrow$61.9 at 1-shot, 89.4$\rightarrow$85.7 at 16-shots). The gains on Caltech and Food101 show that MGT offers a structural cross-modal benefit beyond low-shot initialization, enabling visual nodes to attend to semantic text cues and resolve ambiguities.

\noindent\textbf{Teacher internal design ablations.} Table~\ref{tab:component_ablation} shows that the teacher components are complementary. Removing MGT \textbf{M} causes the largest drop, confirming that \textit{relation-aware cross-modal reasoning} is the main driver (e.g., EuroSAT: 67.4$\rightarrow$61.9, Aircraft: 31.0$\rightarrow$25.6, UCF101: 74.5$\rightarrow$69.9 at 1-shot). Removing \textit{patch–text connections} \textbf{P} also significantly degrades performance, showing that unimodal patch interactions alone are insufficient. In contrast, unimodal Transformer enrichment \textbf{T} and TopK filtering \textbf{F} provide smaller but consistent gains by refining local features and suppressing noise. Overall, \textbf{M}+\textbf{P} contributes the most, while \textbf{T}+\textbf{F} further stabilizes the teacher signal.

\begin{table}[ht]
    \centering
    \scriptsize
    \caption{Internal teacher ablation. \textbf{T:} unimodal Transformer, \textbf{M:} MGT, \textbf{F:} TopK filtering, \textbf{P:} patch–text edges. Cross-modal graph reasoning is the primary contributor (\textbf{M}+\textbf{P}), while \textbf{T} and \textbf{F} offer smaller complementary gains.
    }
    \label{tab:component_ablation}
    \resizebox{\linewidth}{!}{%
    \begin{tabular}{@{}c cccc || ccccc@{}}
        \toprule
        & \textbf{T} & \textbf{M} & \textbf{F} & \textbf{P} & \textbf{1-shot} & \textbf{2-shot} & \textbf{4-shot} & \textbf{8-shot} & \textbf{16-shot} \\
        \midrule
        \multirow{5}{*}{\rotatebox[origin=c]{90}{\textit{EuroSAT}}} 
        & \gmark & \gmark & \gmark & \gmark & \cellcolor{gray!15}\textbf{67.4} & \cellcolor{gray!15}\textbf{74.9} & \cellcolor{gray!15}\textbf{80.3} & \cellcolor{gray!15}\textbf{84.1} & \cellcolor{gray!15}\textbf{89.4} \\
        & \gmark & \gmark & \gmark & \rmark & 64.1 & 71.3 & 76.8 & 81.0 & 86.7 \\
        & \gmark & \gmark & \rmark & \gmark & 63.4 & 72.2 & 79.5 & 83.4 & 86.5 \\
        & \gmark & \rmark & \gmark & \gmark & 61.9 & 71.5 & 77.2 & 80.3 & 85.7 \\
        & \rmark & \gmark & \gmark & \gmark & 66.2 & 72.6 & 78.9 & 81.8 & 87.4 \\
        \midrule
        \multirow{5}{*}{\rotatebox[origin=c]{90}{\textit{Aircraft}}} 
        & \gmark & \gmark & \gmark & \gmark & \cellcolor{gray!15}\textbf{31.0} & \cellcolor{gray!15}\textbf{34.8} & \cellcolor{gray!15}\textbf{38.3} & \cellcolor{gray!15}\textbf{44.2} & \cellcolor{gray!15}\textbf{48.4} \\
        & \gmark & \gmark & \gmark & \rmark & 28.2 & 31.6 & 35.2 & 42.1 & 45.9 \\
        & \gmark & \gmark & \rmark & \gmark & 26.4 & 29.2 & 32.6 & 40.0 & 42.8 \\
        & \gmark & \rmark & \gmark & \gmark & 25.6 & 28.4 & 31.3 & 38.2 & 41.5 \\
        & \rmark & \gmark & \gmark & \gmark & 27.1 & 30.4 & 33.9 & 41.4 & 44.7 \\
        \midrule
        \multirow{5}{*}{\rotatebox[origin=c]{90}{\textit{UCF101}}} 
        & \gmark & \gmark & \gmark & \gmark & \cellcolor{gray!15}\textbf{74.5} & \cellcolor{gray!15}\textbf{77.9} & \cellcolor{gray!15}\textbf{81.7} & \cellcolor{gray!15}\textbf{83.9} & \cellcolor{gray!15}\textbf{84.9} \\
        & \gmark & \gmark & \gmark & \rmark & 71.8 & 74.7 & 78.8 & 79.5 & 81.6 \\
        & \gmark & \gmark & \rmark & \gmark & 71.2 & 74.1 & 78.3 & 78.8 & 80.8 \\
        & \gmark & \rmark & \gmark & \gmark & 69.9 & 72.8 & 76.6 & 76.5 & 78.1 \\
        & \rmark & \gmark & \gmark & \gmark & 73.3 & 76.8 & 80.4 & 83.2 & 83.1 \\
        \bottomrule
    \end{tabular}%
    }
\end{table}

%% file: sec/5_conclusion.tex
\section{Conclusion}
We introduce \textbf{TOGA}, an asymmetric supervision framework for few-shot VLM adaptation that balances efficiency and capability. It leverages a training-only, high-capacity Heterogeneous Image-Patch-Text Graph teacher, employing a Modality-aware Graph Transformer (MGT) to model fine-grained cross-modal relations. This structural knowledge is transferred to a lightweight student adapter via a cache-aware dual-objective strategy, utilizing Focal Loss for teacher regularization. Crucially, this adapter incurs zero extra latency or memory overhead at deployment, as the complex graph teacher is discarded. Our approach establishes new state-of-the-art results across 11 benchmarks, with significant gains in the 1- to 4-shot regimes over global-feature and patch-based baselines. We validate that transferring complex relational reasoning into simple, efficient cache-based prototypes is a highly effective strategy for few-shot VLM adaptation.

\section*{Acknowledgement}

This work was supported by UK Research and Innovation (UKRI) through
the Engineering and Physical Sciences Research Council (EPSRC) grant
ATRACT (\texttt{EP/X028631/1}), the Economic and Social Research Council (ESRC) grant SCAnDi (\texttt{ES/Y010655/1}), and Cancer Research UK (CRUK) grants \texttt{EDDISA-May21\textbackslash100001} and
\texttt{EDDPMA-May22\textbackslash100028}.

%% file: sec/X_suppl.tex
\clearpage
\setcounter{page}{1}
\maketitlesupplementary

\setcounter{section}{0}
\renewcommand{\thesection}{S\arabic{section}}

\setcounter{table}{0}
\renewcommand{\thetable}{S\arabic{table}}

\setcounter{figure}{0}
\renewcommand{\thefigure}{S\arabic{figure}}

\begin{figure*}[!htbp]
  \centering
  \begin{subfigure}[b]{1.0\textwidth}
    \centering
    \includegraphics[width=\textwidth]{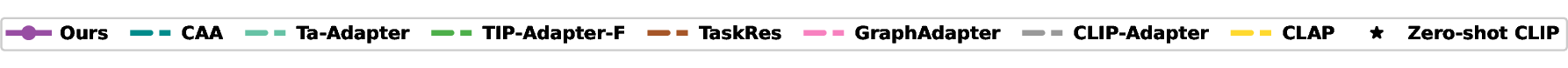}
  \end{subfigure}

  \vspace{-0.2cm} 

  \begin{subfigure}[b]{0.246\textwidth}
    \includegraphics[width=\textwidth]{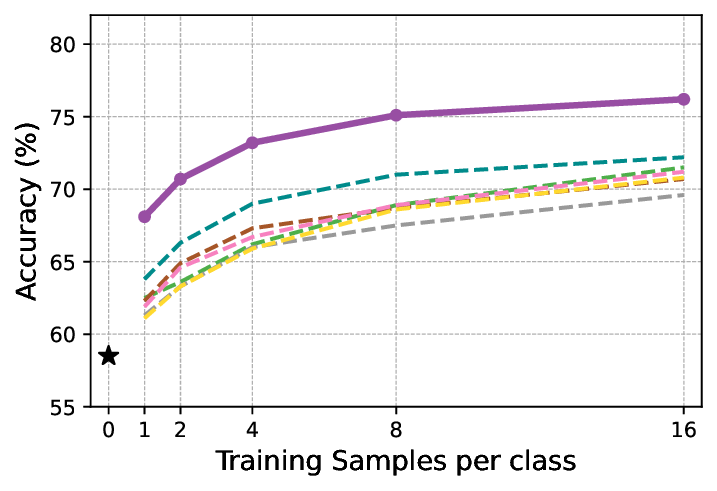}
    \caption{Sun397}
    \label{fig:sun397}
  \end{subfigure}
  \hfill
  \begin{subfigure}[b]{0.246\textwidth}
    \includegraphics[width=\textwidth]{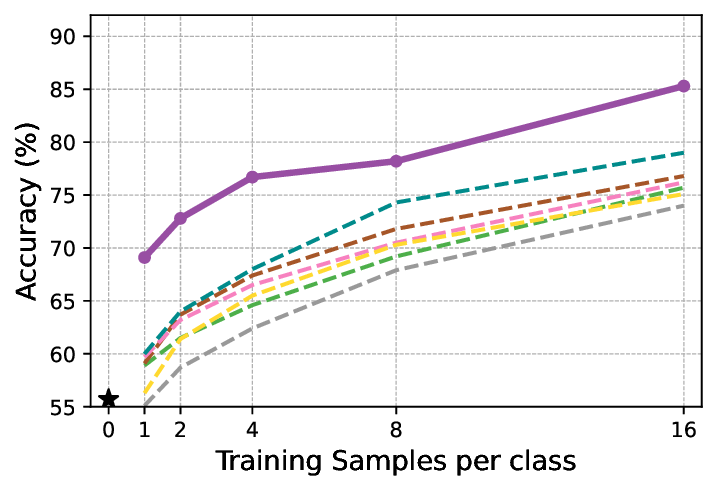}
    \caption{Stanford Cars}
    \label{fig:cars}
  \end{subfigure}
  \hfill
  \begin{subfigure}[b]{0.246\textwidth}
    \includegraphics[width=\textwidth]{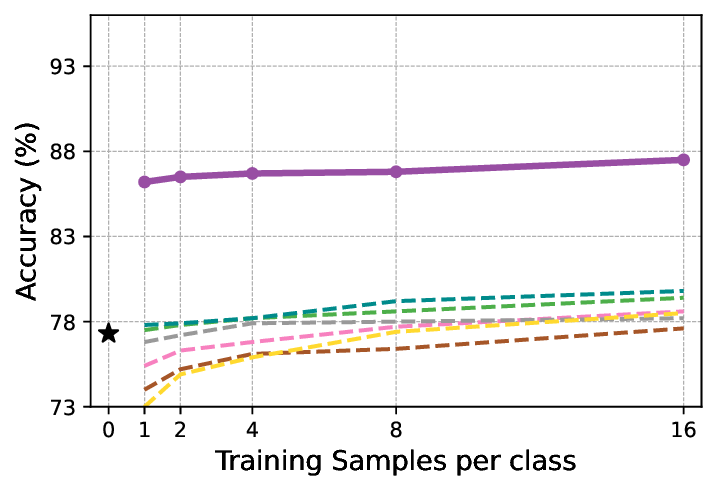}
    \caption{Food101}
    \label{fig:Food101}
  \end{subfigure} 
  \hfill
  \begin{subfigure}[b]{0.246\textwidth}
    \includegraphics[width=\textwidth]{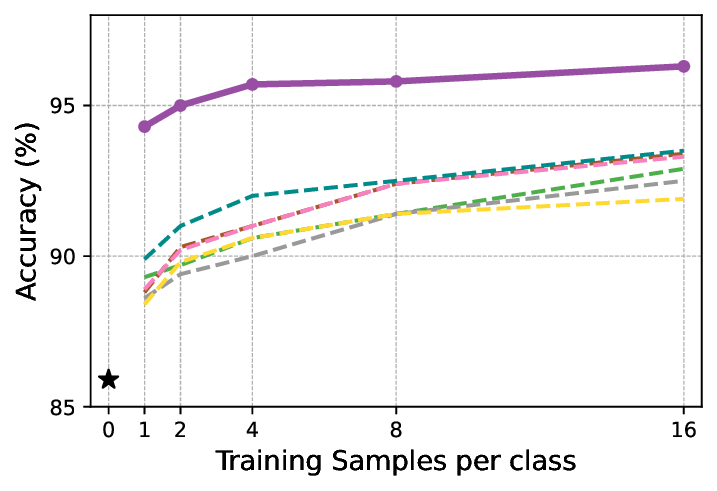}
    \caption{Caltech}
    \label{fig:caltech}
  \end{subfigure} \\
  \hfill
  
  \begin{subfigure}[b]{0.246\textwidth}
    \includegraphics[width=\textwidth]{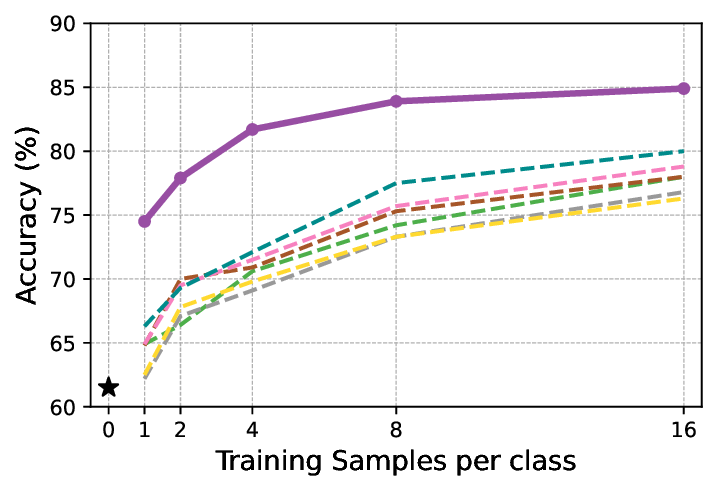}
    \caption{UCF101}
    \label{fig:ucf}
  \end{subfigure} 
  \hfill
  \begin{subfigure}[b]{0.246\textwidth}
    \includegraphics[width=\textwidth]{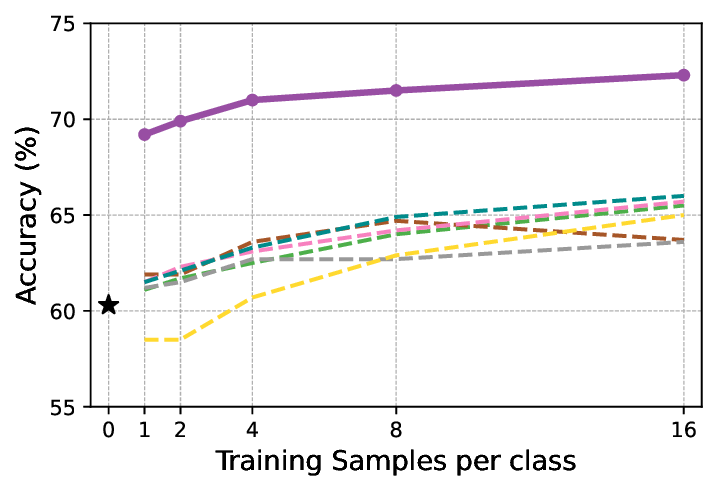}
    \caption{Imagenet}
    \label{fig:imagenet}
  \end{subfigure} 
  \hfill
  \begin{subfigure}[b]{0.246\textwidth}
    \includegraphics[width=\textwidth]{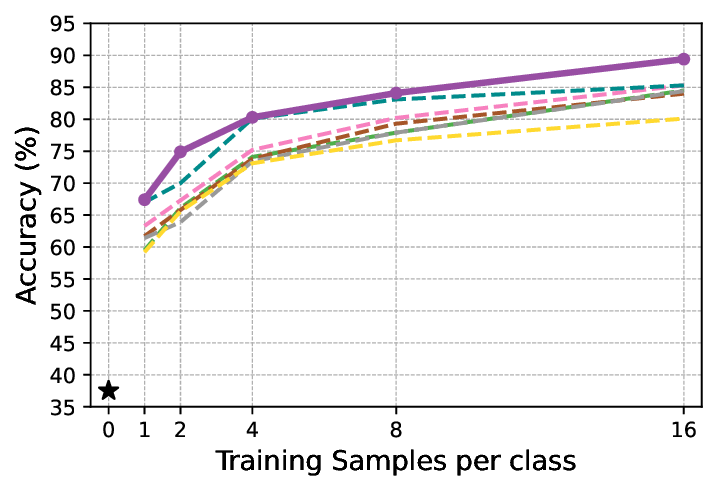}
    \caption{EuroSAT}
    \label{fig:eurosat}
  \end{subfigure} 
  \hfill
  \begin{subfigure}[b]{0.246\textwidth}
    \includegraphics[width=\textwidth]{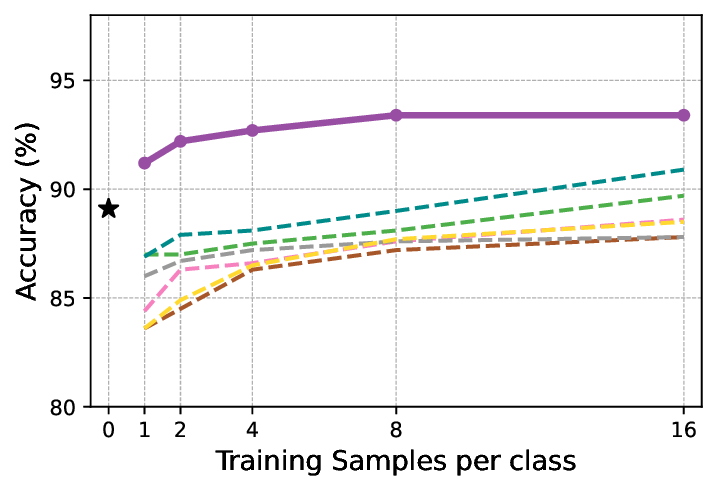}
    \caption{Oxford Pets}
    \label{fig:pets}
  \end{subfigure}%
  
\caption{Performance trajectories of different adapter-based architectures. The curves show the few-shot accuracy of the model on several additional datasets, complementing the main results presented in the paper.}
  \label{fig:fewshot}
\end{figure*}

\noindent To support the findings presented in the main paper, this document details more experiments, ablation studies, and visualizations. Specifically, we cover:

\begin{enumerate}
    \item A comparative analysis of few-shot classification performance against several state-of-the-art (SOTA) prompt-based methods. (Section \ref{sec: A1})
    \item Additional visualization of the dataset for few-shot accuracy curves comparing various adapter-based methods.
    \item t-SNE visualizations comparing the feature distributions of Zero-shot CLIP logits, Tip-Adapter-F logits, and Graph logits. (Section \ref{sec: A2})
    \item More qualitative visualizations of discriminative node filtering. (Section \ref{sec: A3})
    \item Ablation studies evaluating different backbone architectures compared to Tip-Adapter-F \cite{tipadapter}. (Section \ref{sec: A4})
    \item A detailed analysis of computational complexity, feature dimensionality, and model parameters. (Section \ref{sec: A5})
    \item Hyperparameter sensitivity analysis. (Section \ref{sec: A6})
    \item An investigation into the impact of varying the number of multiscale patches/nodes. (Section \ref{sec: A7})
    \item Quantitative visualizations of the Cache and graph model. (Section \ref{sec: A8})
    \item Ablation studies evaluating text nodes in the graph topology and MGT vs. Transformer performance. (Section \ref{sec: A9})
\end{enumerate}

\section{Comparison with SotA Prompt Learning}
\label{sec: A1}
While the main manuscript focuses on adapter-based baselines, we provide here a comprehensive comparison against SOTA prompt learning methods, including static (CoOp \cite{zhou2022learning}), conditional (CoCoOp \cite{zhou2022conditional}, KgCoOp \cite{yao2023visual}), and gradient-aligned (ProGrad \cite{zhu2023prompt}, MaPLe \cite{khattak2023maple}, PLOT++ \cite{chen2022prompt}) approaches. Table \ref{tab:combined_results_prompt} details the performance across 11 benchmarks.

\noindent \textbf{Low-Shot Robustness (1-4 Shots).}
Prompt-based methods often struggle in extreme low-data regimes due to the difficulty of optimizing continuous vectors without overfitting. Our method significantly mitigates this issue.
\begin{itemize}
    \item \textbf{1-Shot:} We achieve an average accuracy of \textbf{72.2\%}, establishing a new SOTA by surpassing the optimal-transport method PLOT++ (70.7\%) by \textbf{+1.5\%} and the multi-modal MaPLe (69.3\%) by \textbf{+2.9\%}.
    \item \textbf{2-Shot:} The gap widens against trajectory-constrained methods like ProGrad (72.4\%) and MaPLe (72.6\%), with our method reaching \textbf{75.0\%}. This indicates that our \textit{Modality-aware Graph Teacher} extracts robust prototypes even from singular examples, whereas prompt gradients often settle into sharp, non-generalizable minima when supervised by sparse data.
    \item \textbf{4-Shot:} This performance advantage is sustained as support size increases. Our method achieves \textbf{77.9\%}, maintaining a clear lead over PLOT++ (76.9\%) and MaPLe (75.8\%). This demonstrates that even as sufficient data becomes available to stabilize prompt optimization, our structure-aware supervision yields superior feature alignment compared to global prompt tuning.
\end{itemize}

\noindent\textbf{High-Shot Scalability and Fine-Grained Tasks (8-16 Shots)}
Our method exhibits sustained superiority in high-shot regimes, achieving an average accuracy of 82.3\% at the 16-shot level, marginally leading the strongest competitor, PLOT++ (82.1\%). This advantage is most pronounced in fine-grained tasks requiring high local discriminative power. For instance, on Aircraft dataset, we reach 48.4\% (matching MaPLe and outperforming ProGrad's 43.0\%), and on Stanford Cars, we achieve 85.3\% (surpassing PLOT++'s 84.5\%). This consistent edge is maintained across the full fine-grained suite, posting 98.3\% on Flowers and a competitive 73.6\% on the challenging task in DTD (outperforming MaPLe's 71.3

\noindent\textbf{Zero-Overhead Inference}
Crucially, while top-performing prompt methods like MaPLe and PLOT++ incur significant overhead due to deep prompt coupling or optimal transport problems, our method distills these performance gains into a static key-value cache. This innovative design allows us to maintain the zero-overhead inference speed of a standard Tip-Adapter-F, effectively outperforming computationally heavier prompt baselines without sacrificing efficiency.

\begin{table*}[!htbp]
\centering
\caption{Comparison of few-shot classification accuracy (\%) on 11 benchmark datasets. We evaluate our method against several SOTA prompt-based approaches. The best performance in each shot-group is marked in \textbf{bold}. Our results are highlighted in gray. \textbf{Dataset abbreviations:} INet (ImageNet), SUN (SUN397), Air (Aircraft), Euro (EuroSAT), Cars (Stanford Cars), Food (Food101), Pets (OxfordPets), Flow (Flowers102), Cal (Caltech101), DTD (Describable Textures), UCF (UCF101).}
\label{tab:combined_results_prompt}
\setlength{\tabcolsep}{5pt} 
\begin{tabular}{@{}l l l r r r r r r r r r r r r@{}}
\toprule
Shots & Method & Venue & INet & SUN & Air & Euro & Cars & Food & Pets & Flow & Cal & DTD & UCF & Avg \\
\midrule

\multirow{1}{*}{0}
& CLIP \textsuperscript{\citep{radford2021clip}} & ICML'22 & 66.7 & 62.6 & 24.7 & 47.5 & 65.3 & 86.1 & 89.1 & 71.4 & 92.9 & 43.6 & 66.7 & 65.1 \\
\midrule

\multirow{8}{*}{1}
& CoOp \textsuperscript{\citep{zhou2022learning}} & IJCV'22 & 68.0 & 67.3 & 26.2 & 50.9 & 67.1 & 82.6 & 90.3 & 72.7 & 93.2 & 50.1 & 70.7 & 67.2 \\
& CoOp \textsuperscript{\citep{zhou2022learning}} & IJCV'22 & 65.7 & 67.0 & 20.8 & 56.4 & 67.5 & 84.3 & 90.2 & 78.3 & 92.5 & 50.1 & 71.2 & 67.6 \\
& CoCoOp \textsuperscript{\citep{zhou2022conditional}} & CVPR'22 & \textbf{69.4} & \textbf{68.7} & 28.1 & 55.4 & 67.6 & 84.9 & \textbf{91.9} & 73.4 & 94.1 & 52.6 & 70.4 & 68.8 \\
& PLOT++ \textsuperscript{\citep{chen2022prompt}} & ICLR'23 & 66.4 & 66.7 & 28.6 & 65.4 & 68.8 & 86.1 & 91.8 & 80.4 & 94.3 & 54.5 & 74.3 & 70.7 \\
& KgCoOp \textsuperscript{\citep{yao2023visual}} & CVPR'23 & 68.9 & 68.4 & 26.8 & 61.9 & 66.7 & \textbf{86.4} & 92.1 & 74.7 & 94.2 & 52.7 & 72.8 & 69.6 \\
& MaPLe \textsuperscript{\citep{khattak2023maple}} & CVPR'23 & 62.6 & 64.7 & 26.7 & 71.8 & 66.6 & 80.5 & 89.1 & 83.3 & 92.5 & 52.1 & 71.8 & 69.3 \\
& ProGrad \textsuperscript{\citep{zhu2023prompt}} & ICCV'23 & 67.0 & 67.0 & 28.8 & 57.0 & 68.2 & 84.9 & 91.4 & 80.9 & 93.5 & 52.8 & 73.3 & 69.5 \\
\cmidrule(lr){2-15}
& \textbf{TOGA (Ours)} & - & \cellcolor{gray!15}69.2 & \cellcolor{gray!15}68.1 & \cellcolor{gray!15}\textbf{31.0} & \cellcolor{gray!15}\textbf{67.4} & \cellcolor{gray!15}\textbf{69.1} & \cellcolor{gray!15}86.2 & \cellcolor{gray!15}91.2 & \cellcolor{gray!15}\textbf{88.2} & \cellcolor{gray!15}\textbf{94.3} & \cellcolor{gray!15}\textbf{55.2} & \cellcolor{gray!15}\textbf{74.5} & \cellcolor{gray!15}\textbf{72.2} \\
\midrule

\multirow{8}{*}{2}
& CoOp (4) \textsuperscript{\citep{zhou2022learning}} & IJCV'22 & 68.7 & 68.0 & 28.1 & 66.2 & 70.5 & 82.6 & 89.9 & 80.9 & 93.0 & 53.7 & 73.5 & 70.5 \\
& CoOp (16) \textsuperscript{\citep{zhou2022learning}} & IJCV'22 & 67.0 & 67.0 & 25.9 & 65.1 & 70.4 & 84.4 & 89.9 & 88.0 & 93.1 & 54.1 & 74.1 & 70.8 \\
& CoCoOp \textsuperscript{\citep{zhou2022conditional}} & CVPR'22 & \textbf{70.1} & 69.4 & 29.3 & 61.8 & 68.4 & 85.9 & 91.9 & 77.8 & 94.4 & 52.3 & 73.4 & 70.4 \\
& PLOT++ \textsuperscript{\citep{chen2022prompt}} & ICLR'23 & 68.2 & 68.0 & 31.1 & 76.8 & \textbf{73.1} & 86.3 & 92.2 & 89.8 & 94.6 & 56.7 & 76.7 & 74.0 \\
& KgCoOp \textsuperscript{\citep{yao2023visual}} & CVPR'23 & 69.6 & 69.6 & 28.0 & 69.2 & 68.2 & \textbf{86.6} & \textbf{92.3} & 79.8 & 94.5 & 55.3 & 74.6 & 71.6 \\
& MaPLe \textsuperscript{\citep{khattak2023maple}} & CVPR'23 & 65.1 & 67.1 & 30.9 & \textbf{78.3} & 71.6 & 81.4 & 90.8 & 88.9 & 93.9 & 55.5 & 74.6 & 72.6 \\
& ProGrad \textsuperscript{\citep{zhu2023prompt}} & ICCV'23 & 69.1 & 69.0 & 31.1 & 66.3 & 72.4 & 84.8 & 91.5 & 87.5 & 93.6 & 56.0 & 75.6 & 72.4 \\
\cmidrule(lr){2-15}
& \textbf{TOGA (Ours)} & - & \cellcolor{gray!15}69.9 & \cellcolor{gray!15}\textbf{70.7} & \cellcolor{gray!15}\textbf{34.8} & \cellcolor{gray!15}74.9 & \cellcolor{gray!15}72.8 & \cellcolor{gray!15}86.5 & \cellcolor{gray!15}92.2 & \cellcolor{gray!15}\textbf{91.9} & \cellcolor{gray!15}\textbf{95.0} & \cellcolor{gray!15}\textbf{58.4} & \cellcolor{gray!15}\textbf{77.9} & \cellcolor{gray!15}\textbf{75.0} \\
\midrule

\multirow{8}{*}{4}
& CoOp \textsuperscript{\citep{zhou2022learning}} & IJCV'22 & 69.7 & 70.6 & 29.7 & 65.8 & 73.4 & 83.5 & 92.3 & 86.6 & 94.5 & 58.5 & 78.1 & 73.0 \\
& CoOp \textsuperscript{\citep{zhou2022learning}} & IJCV'22 & 70.6 & 69.7 & 30.9 & 69.7 & 74.4 & 84.5 & 92.5 & 92.2 & 94.5 & 59.5 & 77.6 & 74.2 \\
& CoCoOp \textsuperscript{\citep{zhou2022conditional}} & CVPR'22 & 70.8 & 70.4 & 30.6 & 61.7 & 69.5 & 86.3 & 92.7 & 81.5 & 94.8 & 55.7 & 75.3 & 71.8 \\
& PLOT++ \textsuperscript{\citep{chen2022prompt}} & ICLR'23 & 70.4 & 71.7 & 35.2 & 83.2 & 76.2 & 86.4 & 92.5 & 92.9 & 95.1 & 62.4 & 79.7 & 76.9 \\
& KgCoOp \textsuperscript{\citep{yao2023visual}} & CVPR'23 & 70.2 & 71.5 & 32.2 & 71.8 & 69.5 & 86.9 & 92.6 & 87.0 & 95.0 & 58.7 & 77.6 & 73.9 \\
& MaPLe \textsuperscript{\citep{khattak2023maple}} & CVPR'23 & 67.7 & 70.6 & 34.8 & \textbf{84.5} & 75.3 & 81.7 & 91.9 & 92.6 & 94.4 & 61.0 & 78.4 & 75.8 \\
& ProGrad \textsuperscript{\citep{zhu2023prompt}} & ICCV'23 & \textbf{71.3} & 71.7 & 34.1 & 69.6 & 75.0 & 85.4 & 92.1 & 91.1 & 94.4 & 59.7 & 77.9 & 73.9 \\
\cmidrule(lr){2-15}
& \textbf{TOGA (Ours)} & - & \cellcolor{gray!15}71.0 & \cellcolor{gray!15}\textbf{73.2} & \cellcolor{gray!15}\textbf{38.3} & \cellcolor{gray!15}80.3 & \cellcolor{gray!15}\textbf{76.7} & \cellcolor{gray!15}\textbf{86.7} & \cellcolor{gray!15}\textbf{92.7} & \cellcolor{gray!15}\textbf{96.4} & \cellcolor{gray!15}\textbf{95.7} & \cellcolor{gray!15}\textbf{64.5} & \cellcolor{gray!15}\textbf{81.7} & \cellcolor{gray!15}\textbf{77.9} \\
\midrule

\multirow{8}{*}{8}
& CoOp (4) \textsuperscript{\citep{zhou2022learning}} & IJCV'22 & 70.8 & 72.4 & 37.0 & 74.7 & 76.8 & 83.3 & 92.1 & 95.0 & 94.7 & 63.7 & 79.8 & 76.4 \\
& CoOp (16) \textsuperscript{\citep{zhou2022learning}} & IJCV'22 & 70.6 & 71.9 & 38.5 & 77.1 & 79.0 & 82.7 & 91.3 & 94.9 & 94.5 & 64.8 & 80.0 & 76.8 \\
& CoCoOp \textsuperscript{\citep{zhou2022conditional}} & CVPR'22 & 70.8 & 71.5 & 32.4 & 69.1 & 70.4 & \textbf{87.0} & 93.3 & 86.3 & 94.9 & 60.1 & 75.9 & 73.8 \\
& PLOT++ \textsuperscript{\citep{chen2022prompt}} & ICLR'23 & 71.3 & 73.9 & 41.4 & \textbf{88.3} & \textbf{81.2} & 86.5 & 93.0 & 95.4 & 95.5 & 66.4 & 82.8 & 79.6 \\
& KgCoOp \textsuperscript{\citep{yao2023visual}} & CVPR'23 & 70.2 & 72.6 & 34.8 & 73.9 & 72.8 & \textbf{87.0} & 93.0 & 91.5 & 95.1 & 65.6 & 80.0 & 76.0 \\
& MaPLe \textsuperscript{\citep{khattak2023maple}} & CVPR'23 & 70.3 & 73.2 & 42.0 & 87.7 & 79.4 & 83.6 & 92.5 & 95.8 & 95.2 & 66.5 & 81.3 & 78.9 \\
& ProGrad \textsuperscript{\citep{zhu2023prompt}} & ICCV'23 & 71.3 & 73.0 & 37.7 & 77.8 & 78.7 & 86.1 & 92.2 & 95.0 & 94.8 & 63.9 & 80.5 & 77.4 \\
\cmidrule(lr){2-15}
& \textbf{TOGA (Ours)} & - & \cellcolor{gray!15}\textbf{71.5} & \cellcolor{gray!15}\textbf{75.1} & \cellcolor{gray!15}\textbf{44.2} & \cellcolor{gray!15}84.1 & \cellcolor{gray!15}78.2 & \cellcolor{gray!15}86.8 & \cellcolor{gray!15}\textbf{93.4} & \cellcolor{gray!15}\textbf{97.3} & \cellcolor{gray!15}\textbf{95.8} & \cellcolor{gray!15}\textbf{69.6} & \cellcolor{gray!15}\textbf{83.9} & \cellcolor{gray!15}\textbf{80.0} \\
\midrule

\multirow{8}{*}{16}
& CoOp \textsuperscript{\citep{zhou2022learning}} & IJCV'22 & 71.5 & 74.6 & 40.1 & 83.5 & 79.1 & 85.1 & 92.4 & 96.4 & 95.5 & 69.2 & 81.9 & 79.0 \\
& CoOp (16) \textsuperscript{\citep{zhou2022learning}} & IJCV'22 & 71.9 & 74.9 & 43.2 & 85.0 & 82.9 & 84.2 & 92.0 & 96.8 & 95.8 & 69.7 & 83.1 & 80.0 \\
& CoCoOp \textsuperscript{\citep{zhou2022conditional}} & CVPR'22 & 71.1 & 72.6 & 33.3 & 73.6 & 72.3 & \textbf{87.4} & 93.4 & 89.1 & 95.1 & 63.7 & 77.2 & 75.3 \\
& PLOT++ \textsuperscript{\citep{chen2022prompt}} & ICLR'23 & 72.6 & 76.0 & 46.7 & 92.0 & 84.5 & 87.1 & \textbf{93.5} & 97.5 & 96.0 & 71.4 & \textbf{85.3} & 82.1 \\
& KgCoOp \textsuperscript{\citep{yao2023visual}} & CVPR'23 & 70.4 & 73.3 & 36.5 & 76.2 & 74.8 & 87.2 & 93.2 & 93.4 & 95.2 & 68.7 & 81.7 & 77.3 \\
& MaPLe \textsuperscript{\citep{khattak2023maple}} & CVPR'23 & \textbf{72.3} & 75.5 & \textbf{48.4} & \textbf{92.3} & 83.5 & 85.3 & 92.8 & 97.0 & 96.0 & 71.3 & 85.0 & 81.8 \\
& ProGrad \textsuperscript{\citep{zhu2023prompt}} & ICCV'23 & 72.1 & 75.1 & 43.0 & 83.6 & 82.9 & 85.8 & 92.8 & 96.6 & 95.9 & 68.8 & 82.7 & 79.9 \\
\cmidrule(lr){2-15}
& \textbf{TOGA (Ours)} & - & \cellcolor{gray!15}\textbf{72.3} & \cellcolor{gray!15}\textbf{76.2} & \cellcolor{gray!15}\textbf{48.4} & \cellcolor{gray!15}89.4 & \cellcolor{gray!15}\textbf{85.3} & \cellcolor{gray!15}87.5 & \cellcolor{gray!15}93.4 & \cellcolor{gray!15}\textbf{98.3} & \cellcolor{gray!15}\textbf{96.3} & \cellcolor{gray!15}\textbf{73.6} & \cellcolor{gray!15}84.9 & \cellcolor{gray!15}\textbf{82.3} \\
\bottomrule
\end{tabular}
\end{table*}

\section{Qualitative Analysis of Embedding Separability via t-SNE}
\label{sec: A2}
To qualitatively validate the effectiveness of our asymmetric supervision, we visualize the logit distributions using t-Distributed Stochastic Neighbor Embedding (t-SNE). We compare the feature manifolds of our method against the baseline Zero-shot CLIP \cite{yang2024clip} and Tip-Adapter-F \cite{tipadapter} in a $16$-shot setting. To ensure a rigorous evaluation of decision boundary refinement, we selected a subset of 8 randomly selected classes per dataset.

\begin{figure*}[!t]
\centering\includegraphics[width=1.0\linewidth]{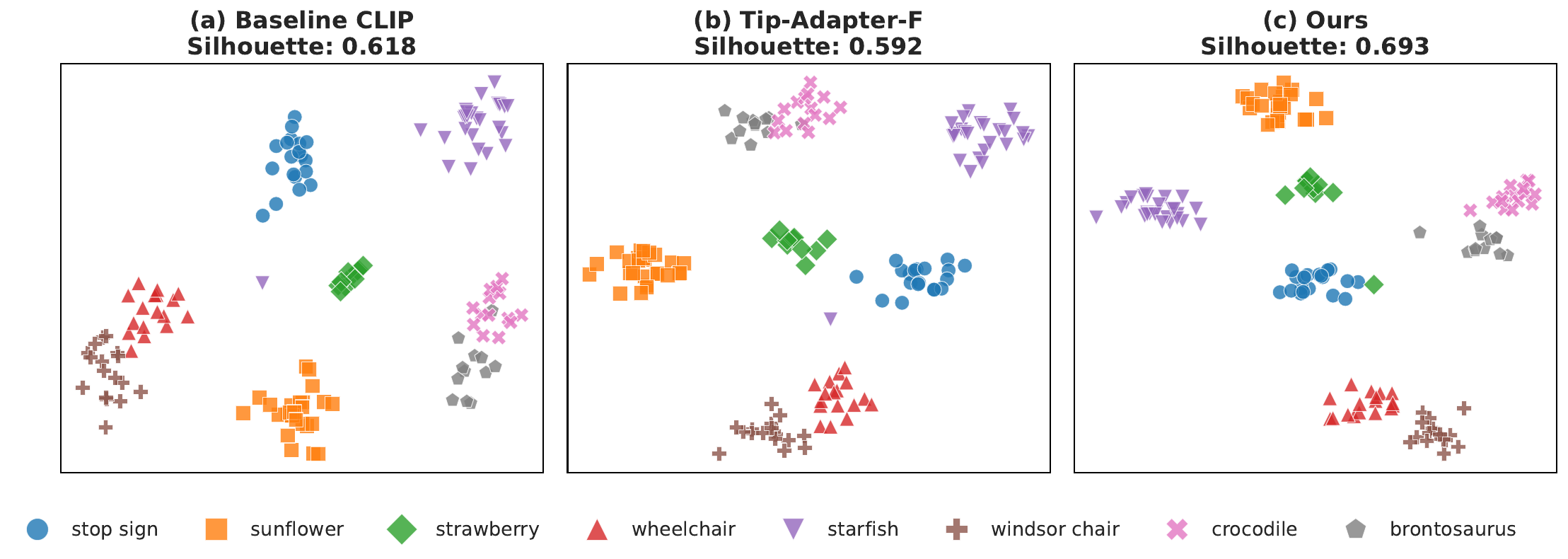}
\caption{t-SNE visualization of logits on the Caltech101 dataset ($16$-shot).
The plots depict the separation of $8$ object categories, ranging from distinct objects like stop sign to complex shapes like windsor chair. (a) Baseline CLIP provides a strong initial separation ($0.618$). (b) Tip-Adapter-F introduces minor variance, slightly reducing the score to $0.592$. (c) Ours further refines the manifold, tightening the clusters for difficult classes (e.g., crocodile and brontosaurus) and achieving the highest distinctiveness with a Silhouette score of $0.693$. This highlights the benefit of our discriminative node filtering in suppressing background noise.}
\label{fig:tsne-cal}
\end{figure*}

\begin{figure*}[!t]
\centering\includegraphics[width=1.0\linewidth]{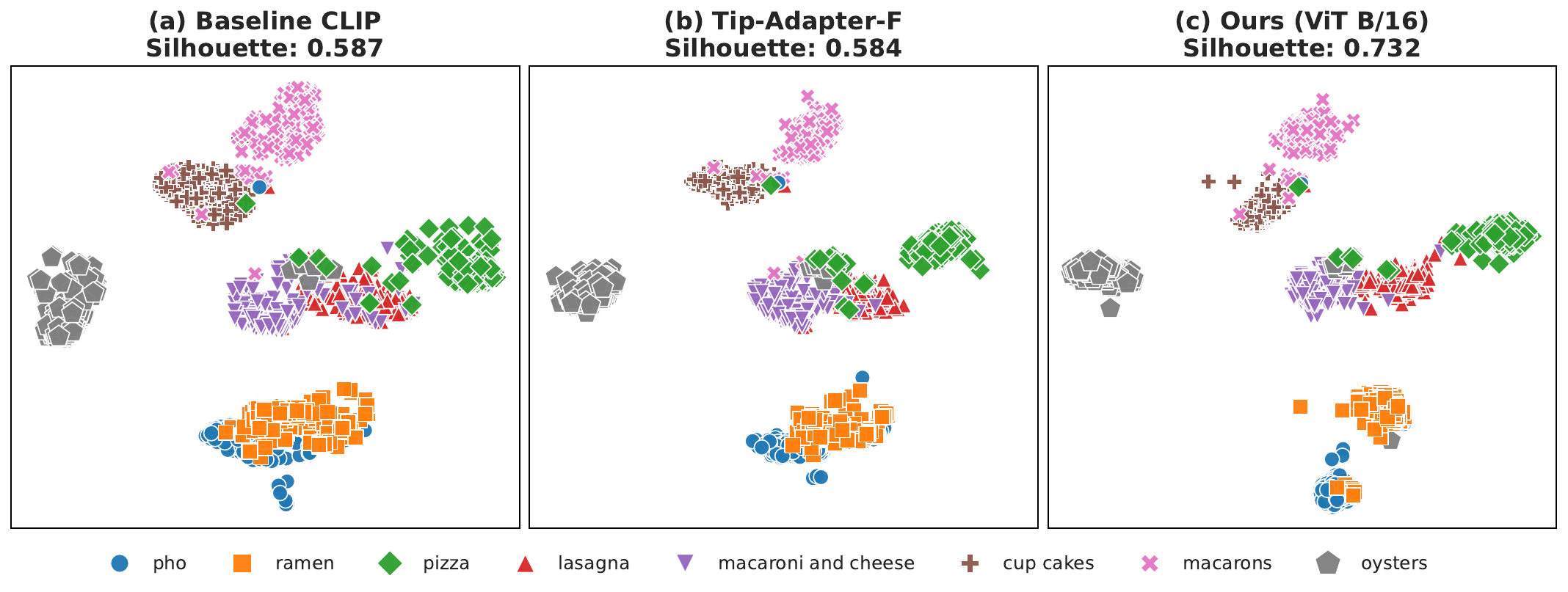}
\caption{This visualization highlights the model's ability to distinguish between fine-grained culinary categories. (a) Baseline CLIP and (b) Tip-Adapter-F show comparable performance ($\approx 0.58$), with noticeable overlap between visually similar classes (e.g., macarons and cup cakes). (c) Ours (ViT-B/16) significantly enhances class purity, leading to superior clustering of features and raising the Silhouette score to $0.732$. This confirms that our MGT-guided distillation effectively transfers local ingredient-level details into the student cache.}
\label{fig:tsne-food}
\end{figure*}

\begin{figure*}[!t]
\centering\includegraphics[width=1.0\linewidth]{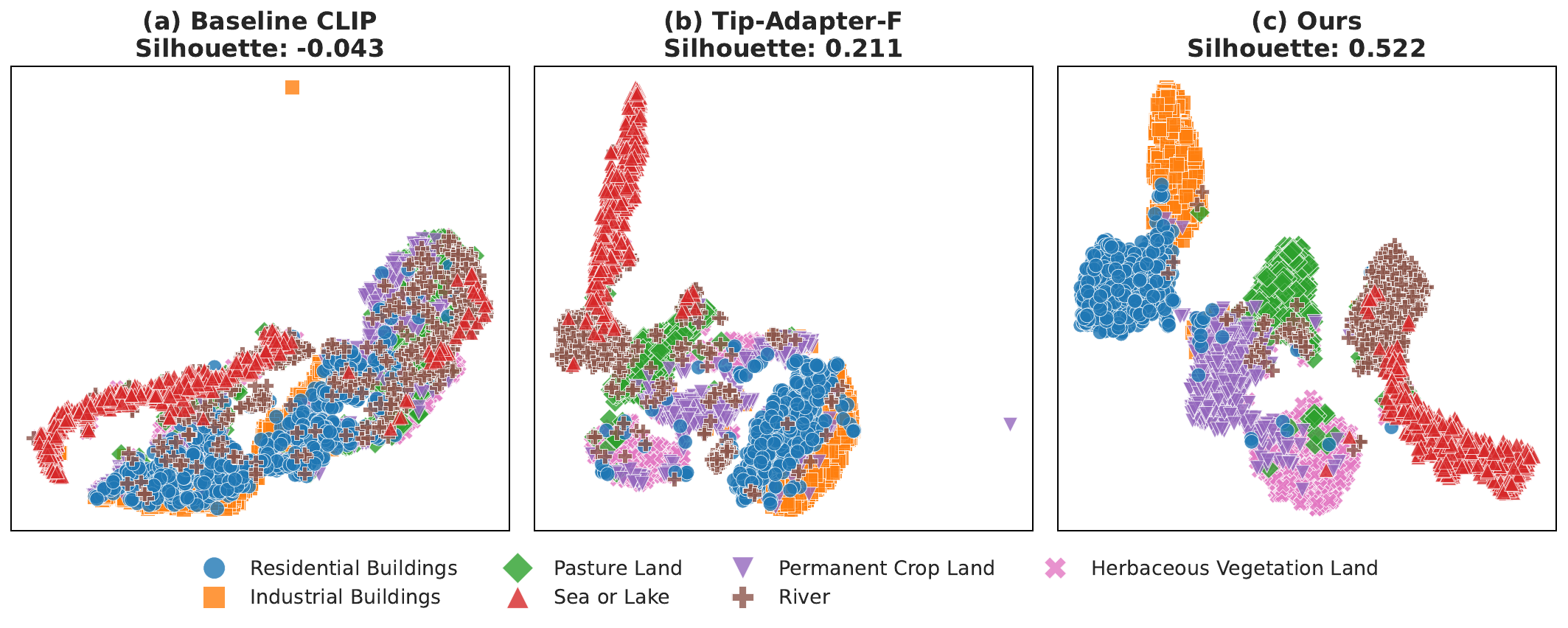}
\caption{t-SNE visualization of logits on the EuroSAT dataset ($16$-shot).
We visualize the embedding space for $8$ randomly selected categories. (a) Baseline CLIP fails to form distinct clusters for texture-heavy classes, resulting in a negative Silhouette score of $-0.043$. (b) Tip-Adapter-F improves separability ($0.211$) but retains ambiguity between vegetation classes. (c) Ours demonstrates that the asymmetric graph supervision effectively disentangles complex spectral signatures, achieving a Silhouette score of $0.522$ and showing clear separation between semantically adjacent classes like Pasture Land and Herbaceous Vegetation Land.}
\label{fig:tsne-euro}
\end{figure*}

The resulting visualizations (Figures \ref{fig:tsne-cal}, \ref{fig:tsne-food}, and \ref{fig:tsne-euro}), illustrate the impact of our Training-Only modality-aware Graph Teacher. While global-feature baselines (CLIP and Tip-Adapter-F) often exhibit diffuse clustering and significant inter-class overlap among semantically similar categories, our method achieves superior intra-class compactness and inter-class separability. This improvement confirms that the fine-grained, cross-modal reasoning performed by the Modality-aware Graph Transformer (MGT) is successfully distilled into the student adapter's key-value cache. The quantitative improvement is evidenced by the Silhouette scores; for example, on the EuroSAT dataset, our approach achieves a score of $0.522$ compared to the baseline's $-0.043$, indicating that our graph-guided teacher forcing effectively rectifies decision boundaries even in domain-shifted remote sensing tasks.

\section{Qualitative Visualization of Discriminative Node Filtering}
\label{sec: A3}
\begin{figure*}[!t]
\centering\includegraphics[width=1.0\linewidth]{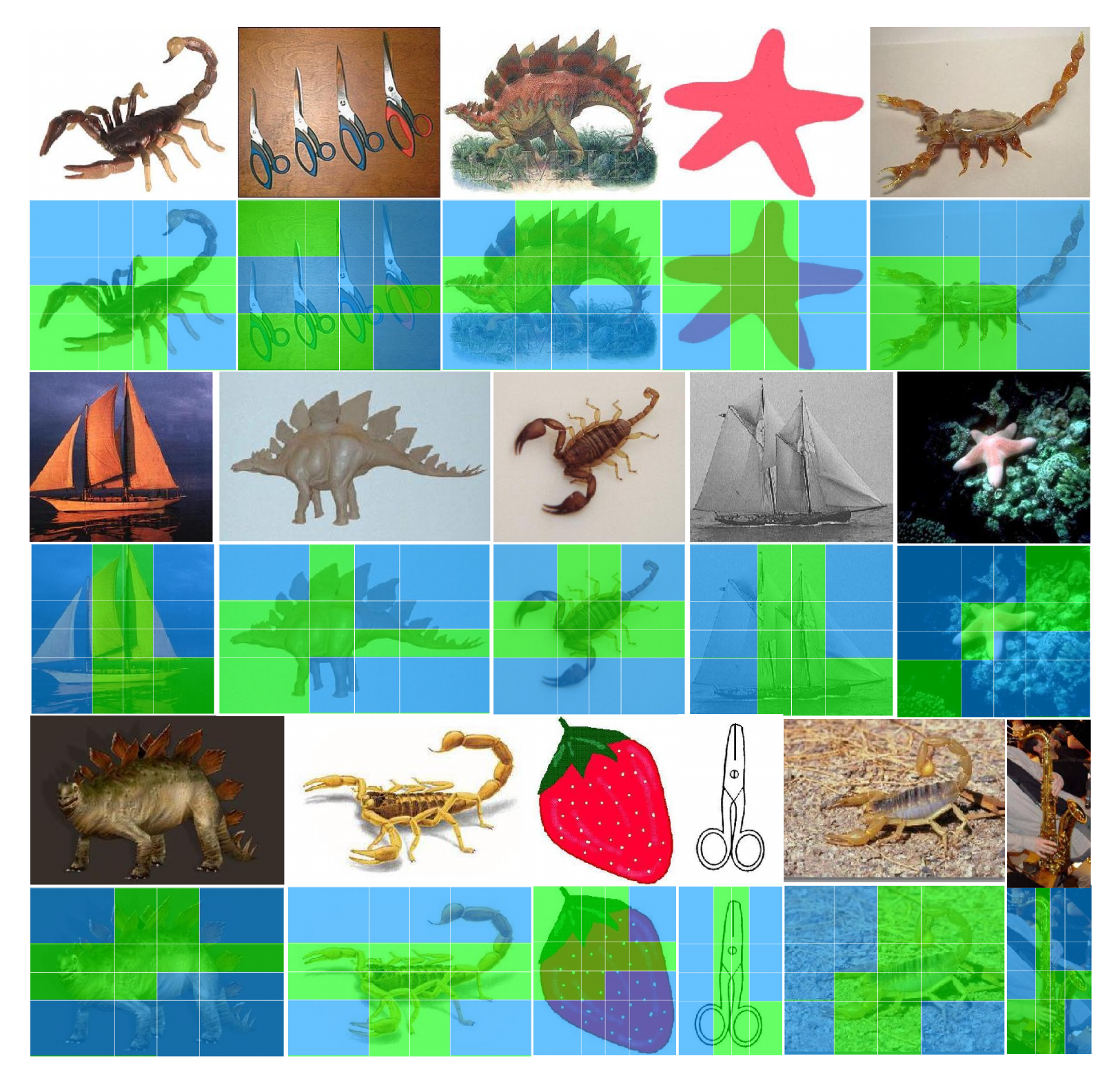}
\caption{\textbf{Extended Qualitative Visualization of Discriminative Node Filtering.} This figure provides additional samples demonstrating the efficacy of our $\text{Top-}\mathbb{N}$ filtering across various Caltech-101 classes. The learned filter successfully identifies and retains high-scoring, discriminative foreground patches (Green) covering key object features (e.g., the segmented body of the scorpion, the silhouette of the scissors, and the spike pattern of the stegosaurus). Concurrently, it suppresses non-informative background regions (Blue). This mechanism ensures that the GNN Teacher's final visual feature is robust against feature dilution, yielding a high-fidelity relational signal for cache supervision.}
\label{fig:kmoreablation}
\end{figure*}

The provided visualization further confirms the necessity of discriminative node filtering to ensure the quality of the teacher's supervisory signal. Our $\text{Top-}\mathbb{N}$ filtering mechanism is designed to mitigate the inherent feature dilution that occurs when aggregating raw patch features, which is especially critical when dealing with object-centric datasets like $\text{Caltech-101}$ that contain diverse object scales and cluttered backgrounds. The visualization (Figure \ref{fig:kmoreablation}) shows that the learned filtering weights, applied after cross-modal reasoning by the GNN Teacher, effectively differentiate between foreground and background elements. Patches covering the central object and defining features, such as the scorpion's body and claws, the scissors' blades and handles, and the distinct outline of the starfish, receive high activation scores (highlighted in {\color{Green}Green}). Conversely, non-informative patches encompassing uniform background areas are suppressed (highlighted in {\color{Blue}Blue}). This selective mechanism consolidates contextualized information into the final visual characteristic ($f_{graph}$) of the teacher's classification head, concentrating primarily on discriminative foreground cues. By generating a cleaner, foreground-centric representation, the filtering enhances the discriminative power of the graph logits, leading to a more robust knowledge transfer during asymmetric supervision.

\section{Backbone Scaling and Efficiency Analysis}
\label{sec: A4}
We validate the scalability of our approach by evaluating its few-shot classification accuracy across four distinct CLIP backbones: $\text{ResNet-101}$ ($\text{RN101}$) and the $\text{ViT-B/32}$, $\text{ViT-B/16}$, and $\text{ViT-L/14}$ Vision Transformer variants, comparing our method against the $\text{Tip-Adapter-F}$ baseline (Base). As shown in Table \ref{tab:redesigned_backbones}, \textbf{TOGA} consistently improves accuracy over the Base model across all architectures and shot settings, demonstrating its broad applicability and robust use of features from diverse encoder structures. As the backbone capacity increases from $\text{ViT-B/32}$ to $\text{ViT-L/14}$, the overall accuracy of our method increases progressively. The highest performance is achieved by the $\text{ViT-L/14}$ model, which obtains an $8$-shot average accuracy of $\mathbf{86.9\%}$. The gains are particularly significant on challenging tasks, such as improving Aircraft accuracy from $53.0\%$ (Base) to $\mathbf{57.0\%}$ (Ours) in $8$-shot with $\text{ViT-L/14}$.

\section{Computational Profile and Asymmetric Cost}
\label{sec: A5}
\begin{figure*}[t]
    \centering
    \includegraphics[width=\linewidth]{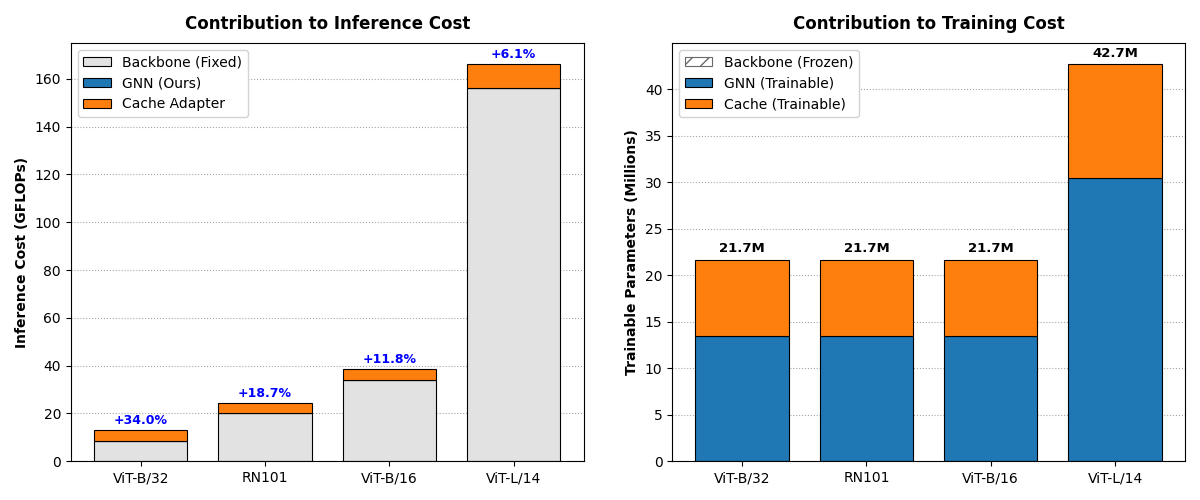}
    \vspace{-0.5cm} 
    \caption{\textbf{Efficiency Profile: Inference vs. Training.} 
    \textbf{Left:} Breakdown of inference computational cost (GFLOPs). Our method (GNN + Cache) introduces only marginal overhead compared to the pre-trained backbone, scaling efficiently to larger models (e.g., only +6.1\% overhead for ViT-L/14). 
    \textbf{Right:} Analysis of trainable parameters. By keeping the massive backbone frozen (grey), our approach requires optimizing only a lightweight set of parameters ($\sim$21M--43M) for the GNN and Adapter, significantly reducing training memory and time costs compared to full fine-tuning.}
    \label{fig:cost_breakdown}
\end{figure*}

The efficiency of our approach comes from its asymmetric supervision design, which minimizes the deployment cost. During training, the total parameter count ($\text{GNN} + \text{Cache}$) remains lightweight; for $\text{ViT-B/16}$, only $21.7$ million parameters are trained, significantly reducing memory and time costs compared to full fine-tuning. Since the GNN teacher module is omitted in inference, the proposed approach does not produce additional latency or memory overhead relative to the $\text{Tip\text{-}Adapter\text{-}F}$ baseline. As shown in Table \ref{tab:efficiency} and Figure \ref{fig:cost_breakdown}, the total cost of inference computation ($\text{GFLOPs}$) for our approach is primarily defined by the fixed cost of the frozen backbone (e.g. $155.7$ $\text{GFLOPs}$ for $\text{ViT-L/14}$) plus a marginal overhead introduced by the lightweight Cache Adapter. This overhead is minimal, increasing the total inference cost by only $6.1\%$ for the high-capacity model $\text{ViT-L/14}$. The Accuracy vs. Efficiency graph confirms this cost-benefit trade-off, showing that high accuracy is achieved with the $\text{ViT-L/14}$ model at its inherent computational cost, demonstrating that graph-guided knowledge transfer effectively maximizes the performance return on the hardware investment.

\begin{table*}[ht!]
\centering
\caption{Comparison of few-shot classification accuracy (\%). The table compares the baseline \textbf{Tip-Adapter-F} (denoted as \textit{Base}) against the \textbf{Proposed} method (denoted as \textit{Ours}) across various CLIP backbones. Distinct blocks represent different shot settings ($K=\{1, 2, 4, 8\}$). Best results are highlighted in \textbf{bold}.}
\label{tab:redesigned_backbones}
\vspace{0.1cm}

\resizebox{\textwidth}{!}{%
\begin{tabular}{@{}cllcccccccccc@{}}
\toprule
\multirow{2}{*}{\textbf{Shots}} & \multirow{2}{*}{\textbf{Backbone}} & \multirow{2}{*}{\textbf{Method}} & \multicolumn{10}{c}{\textbf{Datasets}} \\
\cmidrule(l){4-13} 
 & & & SUN & Air & Euro & Cars & Food & Pets & Flowers & Calt & DTD & UCF \\
\midrule

\multirow{8}{*}{\textbf{1}} 
 & \multirow{2}{*}{ResNet-101} 
   & Base & 60.9 & 19.9 & 38.7 & 64.2 & 80.6 & 87.1 & 64.8 & 91.0 & 43.4 & 63.4 \\
 & & \cellcolor{rowhighlight}\textbf{Ours} & \cellcolor{rowhighlight}63.5 & \cellcolor{rowhighlight}22.7 & \cellcolor{rowhighlight}62.1 & \cellcolor{rowhighlight}66.1 & \cellcolor{rowhighlight}81.0 & \cellcolor{rowhighlight}87.8 & \cellcolor{rowhighlight}81.1 & \cellcolor{rowhighlight}91.9 & \cellcolor{rowhighlight}50.5 & \cellcolor{rowhighlight}68.0 \\
 & \multirow{2}{*}{ViT-B/32}   
   & Base & 63.8 & 20.8 & 49.9 & 61.1 & 80.3 & 87.2 & 67.8 & 91.5 & 48.2 & 65.9 \\
 & & \cellcolor{rowhighlight}\textbf{Ours} & \cellcolor{rowhighlight}66.2 & \cellcolor{rowhighlight}24.3 & \cellcolor{rowhighlight}60.8 & \cellcolor{rowhighlight}62.9 & \cellcolor{rowhighlight}81.0 & \cellcolor{rowhighlight}88.5 & \cellcolor{rowhighlight}84.5 & \cellcolor{rowhighlight}92.6 & \cellcolor{rowhighlight}52.5 & \cellcolor{rowhighlight}70.2 \\
 & \multirow{2}{*}{ViT-B/16}   
   & Base & 65.0 & 25.9 & 53.4 & 67.1 & 85.0 & 89.2 & 69.7 & 93.6 & 47.1 & 70.9 \\
 & & \cellcolor{rowhighlight}\textbf{Ours} & \cellcolor{rowhighlight}68.1 & \cellcolor{rowhighlight}31.0 & \cellcolor{rowhighlight}67.4 & \cellcolor{rowhighlight}69.1 & \cellcolor{rowhighlight}86.2 & \cellcolor{rowhighlight}91.2 & \cellcolor{rowhighlight}88.2 & \cellcolor{rowhighlight}94.3 & \cellcolor{rowhighlight}55.2 & \cellcolor{rowhighlight}74.5 \\
 & \multirow{2}{*}{ViT-L/14}   
   & Base & 70.7 & 37.5 & 65.3 & 79.1 & \textbf{92.0} & 93.1 & 86.1 & 93.8 & 57.0 & 78.5 \\
 & & \cellcolor{rowhighlight}\textbf{Ours} & \cellcolor{rowhighlight}\textbf{75.5} & \cellcolor{rowhighlight}\textbf{43.4} & \cellcolor{rowhighlight}\textbf{71.0} & \cellcolor{rowhighlight}\textbf{93.1} & \cellcolor{rowhighlight}88.4 & \cellcolor{rowhighlight}\textbf{93.7} & \cellcolor{rowhighlight}\textbf{92.1} & \cellcolor{rowhighlight}\textbf{97.4} & \cellcolor{rowhighlight}\textbf{66.5} & \cellcolor{rowhighlight}\textbf{78.8} \\
\midrule \addlinespace

\multirow{8}{*}{\textbf{2}} 
 & \multirow{2}{*}{ResNet-101} 
   & Base & 63.7 & 20.4 & 41.3 & 66.2 & 80.0 & 87.2 & 71.9 & 92.0 & 47.8 & 67.4 \\
 & & \cellcolor{rowhighlight}\textbf{Ours} & \cellcolor{rowhighlight}66.5 & \cellcolor{rowhighlight}25.0 & \cellcolor{rowhighlight}58.8 & \cellcolor{rowhighlight}69.8 & \cellcolor{rowhighlight}81.4 & \cellcolor{rowhighlight}88.8 & \cellcolor{rowhighlight}87.7 & \cellcolor{rowhighlight}93.0 & \cellcolor{rowhighlight}53.9 & \cellcolor{rowhighlight}72.6 \\
 & \multirow{2}{*}{ViT-B/32}   
   & Base & 66.4 & 20.6 & 52.6 & 63.0 & 80.2 & 88.0 & 74.9 & 92.7 & 51.7 & 68.6 \\
 & & \cellcolor{rowhighlight}\textbf{Ours} & \cellcolor{rowhighlight}68.4 & \cellcolor{rowhighlight}26.3 & \cellcolor{rowhighlight}70.1 & \cellcolor{rowhighlight}65.8 & \cellcolor{rowhighlight}81.0 & \cellcolor{rowhighlight}89.3 & \cellcolor{rowhighlight}88.2 & \cellcolor{rowhighlight}95.0 & \cellcolor{rowhighlight}56.4 & \cellcolor{rowhighlight}73.5 \\
 & \multirow{2}{*}{ViT-B/16}   
   & Base & 68.2 & 28.7 & 55.6 & 69.4 & 85.0 & 90.3 & 82.1 & 94.0 & 51.0 & 74.6 \\
 & & \cellcolor{rowhighlight}\textbf{Ours} & \cellcolor{rowhighlight}70.7 & \cellcolor{rowhighlight}34.8 & \cellcolor{rowhighlight}74.9 & \cellcolor{rowhighlight}72.8 & \cellcolor{rowhighlight}86.5 & \cellcolor{rowhighlight}92.2 & \cellcolor{rowhighlight}91.9 & \cellcolor{rowhighlight}95.0 & \cellcolor{rowhighlight}58.4 & \cellcolor{rowhighlight}77.9 \\
 & \multirow{2}{*}{ViT-L/14}   
   & Base & 73.3 & 37.2 & 70.1 & 80.2 & \textbf{92.1} & 93.4 & 88.8 & 95.4 & 61.1 & 81.0 \\
 & & \cellcolor{rowhighlight}\textbf{Ours} & \cellcolor{rowhighlight}\textbf{76.6} & \cellcolor{rowhighlight}\textbf{49.0} & \cellcolor{rowhighlight}\textbf{75.5} & \cellcolor{rowhighlight}\textbf{93.6} & \cellcolor{rowhighlight}88.6 & \cellcolor{rowhighlight}\textbf{94.1} & \cellcolor{rowhighlight}\textbf{95.2} & \cellcolor{rowhighlight}\textbf{97.0} & \cellcolor{rowhighlight}\textbf{67.0} & \cellcolor{rowhighlight}\textbf{81.8} \\
\midrule \addlinespace

\multirow{8}{*}{\textbf{4}} 
 & \multirow{2}{*}{ResNet-101} 
   & Base & 67.8 & 23.5 & 59.9 & 71.6 & 80.3 & 88.4 & 85.5 & 92.5 & 58.2 & 73.0 \\
 & & \cellcolor{rowhighlight}\textbf{Ours} & \cellcolor{rowhighlight}68.9 & \cellcolor{rowhighlight}29.8 & \cellcolor{rowhighlight}77.0 & \cellcolor{rowhighlight}73.4 & \cellcolor{rowhighlight}81.3 & \cellcolor{rowhighlight}90.0 & \cellcolor{rowhighlight}92.5 & \cellcolor{rowhighlight}93.4 & \cellcolor{rowhighlight}62.0 & \cellcolor{rowhighlight}75.2 \\
 & \multirow{2}{*}{ViT-B/32}   
   & Base & 69.3 & 25.7 & 63.4 & 67.0 & 80.5 & 88.7 & 86.5 & 94.5 & 57.8 & 74.6 \\
 & & \cellcolor{rowhighlight}\textbf{Ours} & \cellcolor{rowhighlight}70.2 & \cellcolor{rowhighlight}29.1 & \cellcolor{rowhighlight}79.8 & \cellcolor{rowhighlight}69.4 & \cellcolor{rowhighlight}81.3 & \cellcolor{rowhighlight}89.8 & \cellcolor{rowhighlight}92.1 & \cellcolor{rowhighlight}94.8 & \cellcolor{rowhighlight}61.4 & \cellcolor{rowhighlight}77.0 \\
 & \multirow{2}{*}{ViT-B/16}   
   & Base & 70.6 & 33.9 & 67.6 & 74.2 & 85.6 & 91.0 & 92.4 & 95.2 & 61.6 & 79.2 \\
 & & \cellcolor{rowhighlight}\textbf{Ours} & \cellcolor{rowhighlight}73.2 & \cellcolor{rowhighlight}38.3 & \cellcolor{rowhighlight}80.3 & \cellcolor{rowhighlight}76.7 & \cellcolor{rowhighlight}86.7 & \cellcolor{rowhighlight}92.7 & \cellcolor{rowhighlight}96.4 & \cellcolor{rowhighlight}95.7 & \cellcolor{rowhighlight}64.5 & \cellcolor{rowhighlight}81.7 \\ 
 & \multirow{2}{*}{ViT-L/14}   
   & Base & 76.2 & 45.9 & 76.1 & 84.0 & \textbf{92.1} & \textbf{94.9} & 96.1 & 96.3 & 68.3 & 83.1 \\
& & \cellcolor{rowhighlight}\textbf{Ours} & \cellcolor{rowhighlight}\textbf{78.2} & \cellcolor{rowhighlight}\textbf{52.0} & \cellcolor{rowhighlight}\textbf{83.2} & \cellcolor{rowhighlight}\textbf{93.5} & \cellcolor{rowhighlight}90.5 & \cellcolor{rowhighlight}94.4 & \cellcolor{rowhighlight}\textbf{97.7} & \cellcolor{rowhighlight}\textbf{97.8} & \cellcolor{rowhighlight}\textbf{73.6} & \cellcolor{rowhighlight}\textbf{85.8} \\
\midrule \addlinespace

\multirow{8}{*}{\textbf{8}} 
 & \multirow{2}{*}{ResNet-101} 
   & Base & 69.8 & 29.2 & 67.2 & 75.1 & 81.0 & 89.0 & 91.4 & 93.1 & 64.2 & 77.1 \\
 & & \cellcolor{rowhighlight}\textbf{Ours} & \cellcolor{rowhighlight}70.9 & \cellcolor{rowhighlight}36.1 & \cellcolor{rowhighlight}83.0 & \cellcolor{rowhighlight}75.9 & \cellcolor{rowhighlight}81.7 & \cellcolor{rowhighlight}90.2 & \cellcolor{rowhighlight}95.1 & \cellcolor{rowhighlight}93.9 & \cellcolor{rowhighlight}65.4 & \cellcolor{rowhighlight}78.5 \\
 & \multirow{2}{*}{ViT-B/32}   
   & Base & 69.9 & 30.6 & 74.2 & 71.8 & 80.8 & 88.9 & 92.0 & 94.7 & 64.3 & 78.1 \\
 & & \cellcolor{rowhighlight}\textbf{Ours} & \cellcolor{rowhighlight}72.5 & \cellcolor{rowhighlight}35.4 & \cellcolor{rowhighlight}82.6 & \cellcolor{rowhighlight}73.1 & \cellcolor{rowhighlight}81.3 & \cellcolor{rowhighlight}89.7 & \cellcolor{rowhighlight}95.2 & \cellcolor{rowhighlight}95.1 & \cellcolor{rowhighlight}65.9 & \cellcolor{rowhighlight}80.2 \\
 & \multirow{2}{*}{ViT-B/16}   
   & Base & 73.1 & 38.8 & 77.4 & 77.1 & 86.7 & 92.0 & 95.6 & 95.3 & 67.7 & 81.3 \\
 & & \cellcolor{rowhighlight}\textbf{Ours} & \cellcolor{rowhighlight}75.1 & \cellcolor{rowhighlight}44.2 & \cellcolor{rowhighlight}84.1 & \cellcolor{rowhighlight}78.2 & \cellcolor{rowhighlight}86.8 & \cellcolor{rowhighlight}93.4 & \cellcolor{rowhighlight}97.3 & \cellcolor{rowhighlight}95.8 & \cellcolor{rowhighlight}69.6 & \cellcolor{rowhighlight}83.9 \\
 & \multirow{2}{*}{ViT-L/14}   
   & Base & 78.5 & 53.0 & 78.9 & 86.5 & \textbf{92.3} & 94.5 & 97.6 & 97.2 & 72.0 & 86.2 \\
 & & \cellcolor{rowhighlight}\textbf{Ours} & \cellcolor{rowhighlight}\textbf{79.5} & \cellcolor{rowhighlight}\textbf{57.0} & \cellcolor{rowhighlight}\textbf{83.6} & \cellcolor{rowhighlight}\textbf{94.3} & \cellcolor{rowhighlight}90.7 & \cellcolor{rowhighlight}\textbf{94.6} & \cellcolor{rowhighlight}\textbf{98.2} & \cellcolor{rowhighlight}\textbf{97.7} & \cellcolor{rowhighlight}\textbf{76.0} & \cellcolor{rowhighlight}\textbf{86.9} \\

\bottomrule
\end{tabular}
}
\end{table*}

\begin{figure*}[t]
    \centering
    \includegraphics[width=0.81\linewidth]{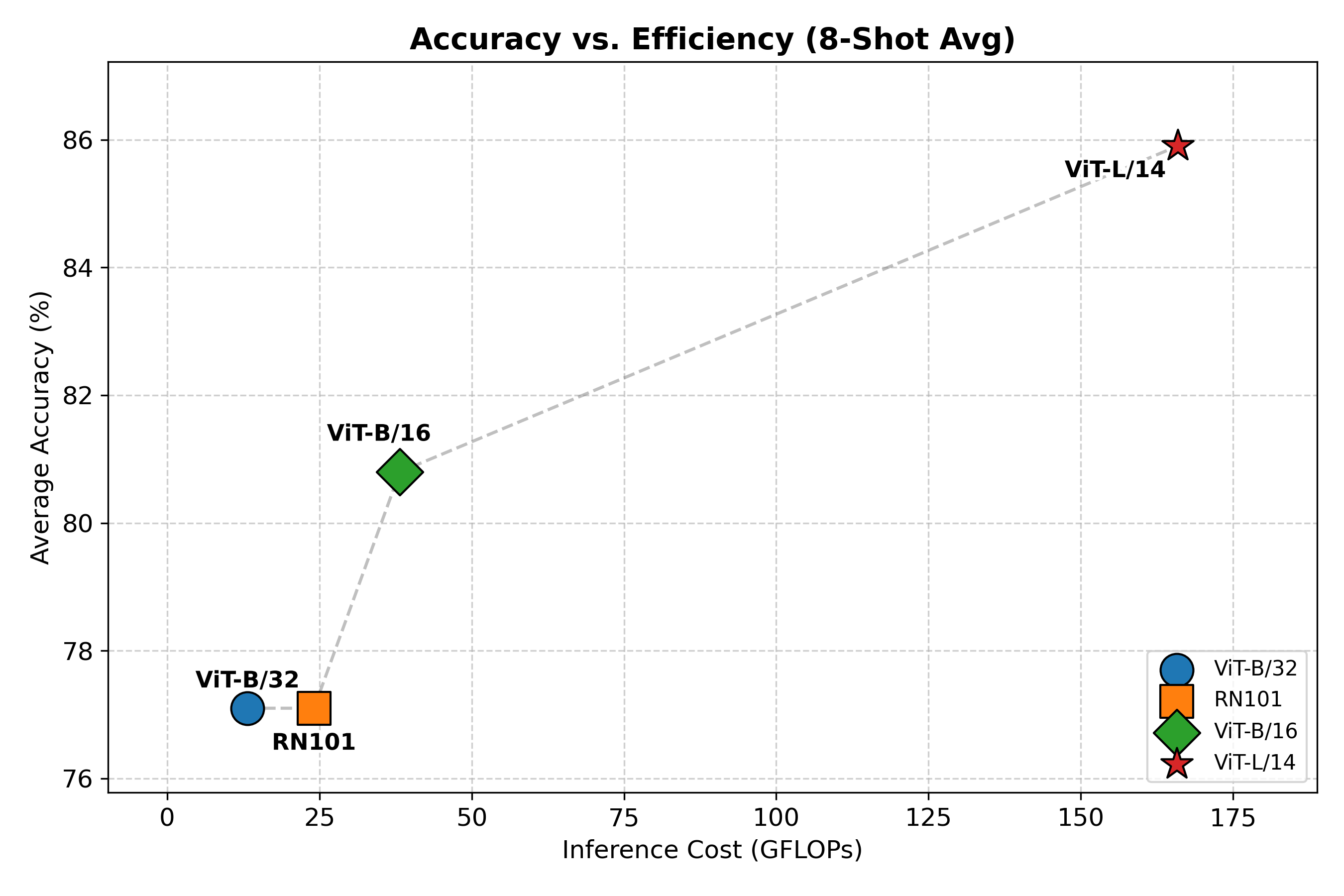} 
    \vspace{-0.2cm} 
    \caption{\textbf{Accuracy vs. Efficiency Trade-off (8-Shot Average).} 
    We compare the total inference cost (GFLOPs) against average few-shot accuracy across 10 benchmark datasets.
    Our method demonstrates \textbf{strong scaling efficiency}: while the transition from ViT-B/16 to ViT-L/14 incurs higher computational cost (due to the larger backbone resolution and depth), it yields a massive \textbf{+5.1\% performance gain}, reaching a state-of-the-art average accuracy of 85.9\%. 
    This highlights that our GNN-based adaptation module effectively leverages the capacity of stronger backbones without hitting a saturation point.}
    \label{fig:acc_eff_tradeoff}
\end{figure*}

\begin{table*}[t]
\centering
\caption{\textbf{Efficiency Analysis.} We report GFLOPs (billions) and Parameters (millions) for the Backbone, our GNN module, and the Cache Adapter. Note that our GNN introduces marginal overhead compared to the backbone.}
\label{tab:efficiency}
\resizebox{0.9\textwidth}{!}{%
\begin{tabular}{l c c | cc | cc | cc}
\toprule
\multirow{2}{*}{\textbf{Backbone}} & \multirow{2}{*}{\textbf{Res.}} & \multirow{2}{*}{\textbf{Dim}} & \multicolumn{2}{c|}{\textbf{Backbone}} & \multicolumn{2}{c|}{\textbf{GNN (Ours)}} & \multicolumn{2}{c}{\textbf{Cache}} \\
 & & & \textbf{GFLOPs} & \textbf{Param} & \textbf{GFLOPs} & \textbf{Param} & \textbf{GFLOPs} & \textbf{Param} \\
\midrule
RN101 & 224 & 512 & 19.6 & 56.3 & 4.5 & 13.5 & 0.02 & 8.2 \\
ViT-B/32 & 224 & 512 & 8.7 & 87.9 & 4.5 & 13.5 & 0.02 & 8.2 \\
ViT-B/16 & 224 & 512 & 33.7 & 86.2 & 4.5 & 13.5 & 0.02 & 8.2 \\
ViT-L/14 & 224 & 768 & 155.7 & 304.0 & 10.2 & 30.4 & 0.03 & 12.3 \\
\bottomrule
\end{tabular}%
}
\end{table*}

\begin{figure*}[t]
    \centering
    \begin{subfigure}[b]{\textwidth}
        \centering
        \includegraphics[width=0.98\linewidth]{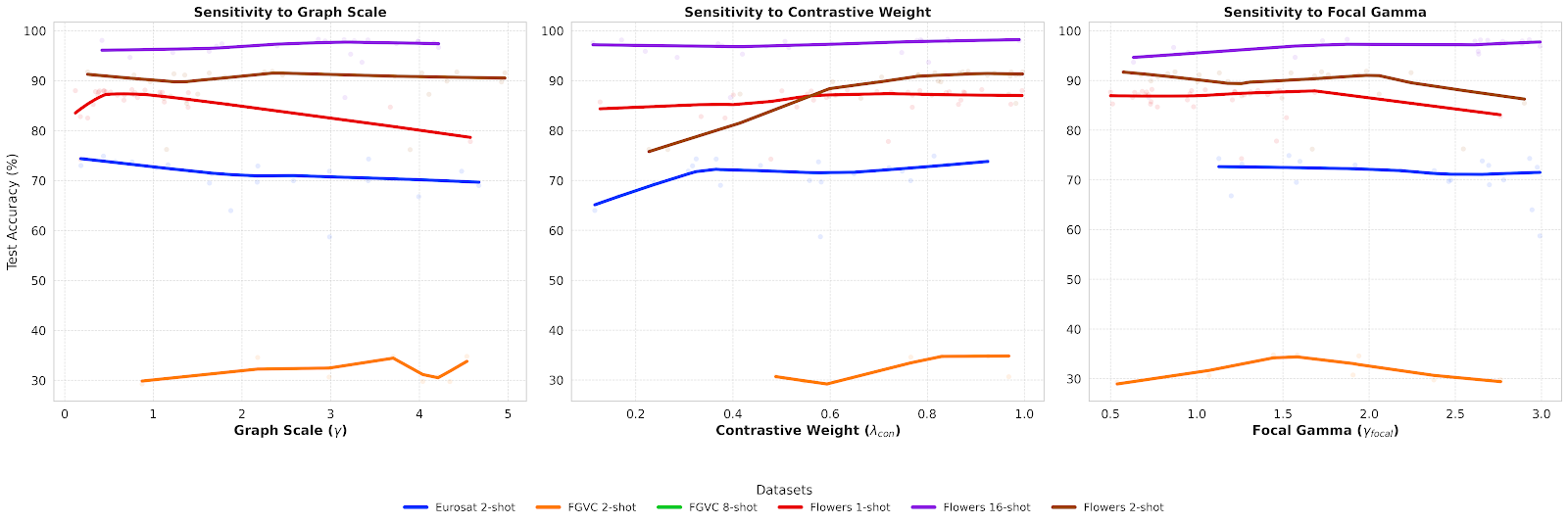} 
        \caption{\textbf{Comprehensive Sensitivity Trends.} LOWESS smoothed curves show the impact of hyperparameters across 6 datasets. Note the varying slopes for $\lambda_{con}$ depending on shot count.}
        \label{fig:trends}
    \end{subfigure}
    \vspace{0.5em}
    
    \begin{subfigure}[b]{\textwidth}
        \centering
        \includegraphics[width=0.98\linewidth]{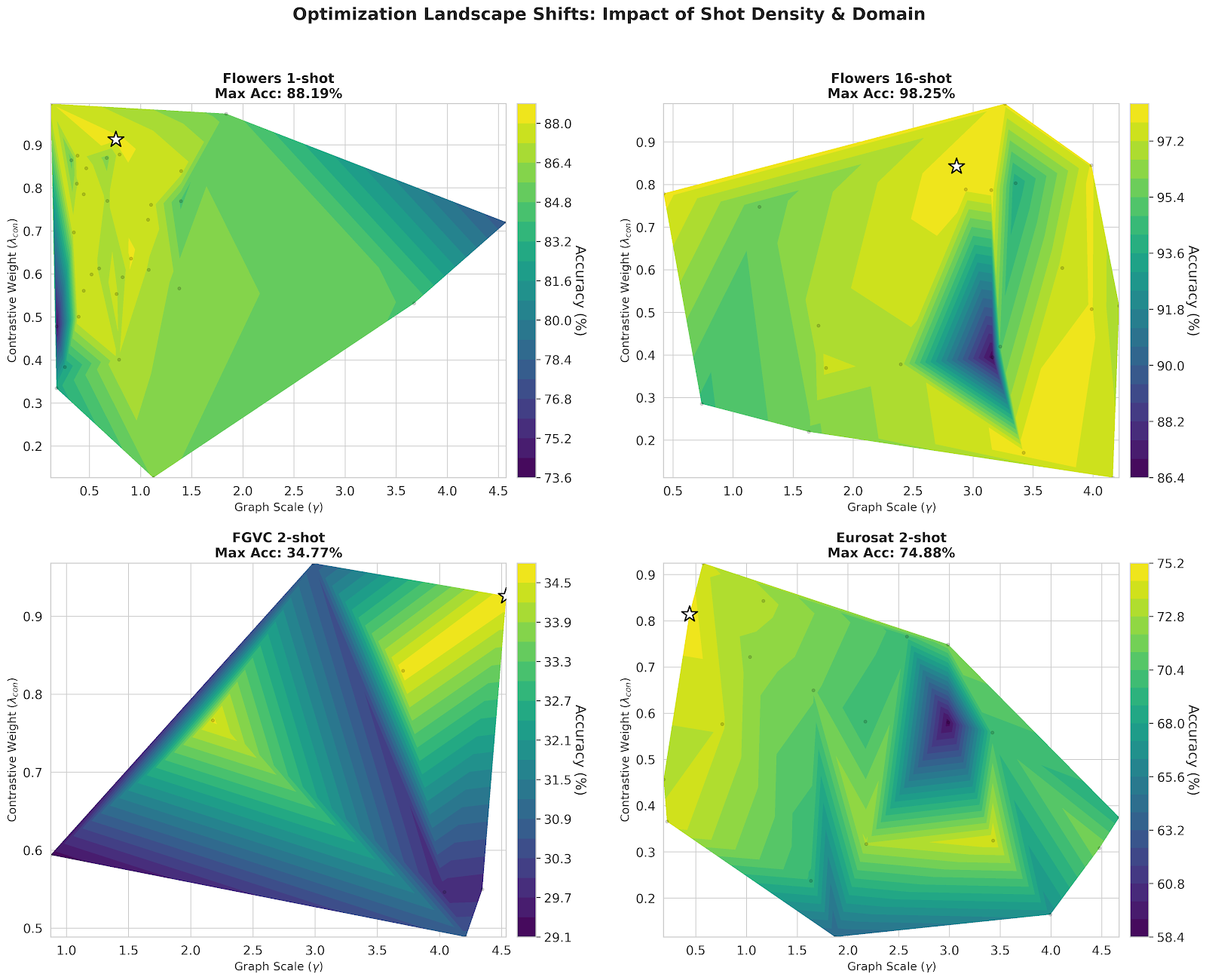}
        \caption{\textbf{Landscape Evolution.} We compare the optimization landscape ($\gamma$ vs. $\lambda_{con}$) across different shots and domains. High-shot settings (e.g., Flowers 16-shot) and distinct domains (Eurosat) exhibit different optimality regions compared to low-shot natural images.}
        \label{fig:landscape}
    \end{subfigure}
    \caption{\textbf{Hyperparameter Sensitivity Analysis.} We visualize the impact of key hyperparameters on test accuracy, highlighting the necessity of our validation-based tuning protocol.}
    \label{fig:sensitivity}
\end{figure*}

\section{Reproducibility and Hyperparameter Tuning}
\label{sec: A6}

We are committed to the full reproducibility of our results. In this section, we detail the exact experimental protocols, code availability, and hyperparameter tuning strategies that ensure our method's performance is stable and reproducible across diverse visual domains.

\subsection{Unified Network Architecture}
To demonstrate that our performance gains stem from topological reasoning rather than architecture engineering, we utilize a \textbf{strictly unified model architecture} across all datasets, ranging from fine-grained Flowers to satellite imagery (Eurosat). We do not alter the model depth or width per task. Specifically, we first employ a 3-layer Transformer encoder with a feed-forward multiplier of $2\times$ to refine the initial features. This is followed by our Modality-aware Graph Transformer (MGT) module with $L=3$ layers and $H=16$ attention heads. The final feature dimension is fixed at $D=512$ to align perfectly with the CLIP ViT-B/16 backbone.

\subsection{Validation Protocol and Search Space}
\label{subsec:search_protocol}
Consistent with standard few-shot learning protocols \cite{tipadapter, zhou2022learning}, we strictly utilize the provided \textbf{few-shot validation set} for hyperparameter selection and explicitly avoid tuning on the test set. We acknowledge that different visual domains require different optimization dynamics. For instance, fine-grained tasks (FGVC) often require stronger contrastive regularization ($\lambda_{con}$) to separate visually similar classes, whereas satellite domains (Eurosat) may require distinct graph scaling ($\gamma$) due to the large semantic shift from the pre-training data. Therefore, rather than claiming a single scalar value works for all domains, we perform a grid search on the validation set within the ranges specified in Table~\ref{tab:search_space}.

\subsection{Sensitivity and Landscape Analysis}
To verify the stability of our method, we analyze test accuracy trends across random hyperparameter sweeps on six dataset configurations, including Flowers (1/2/16-shot), Aircraft (2/8-shot), and Eurosat (2-shot). As visualized in Figure~\ref{fig:sensitivity}, we observe three critical technical insights:

\begin{enumerate}
    \item \textbf{Robustness to Hardness Focus ($\gamma_{focal}$):} The performance curves for the Focal Loss parameter $\gamma_{focal}$ (Figure~\ref{fig:trends}, right) remain relatively flat across all datasets. This indicates that while hard-negative mining is beneficial, the method is optimization-stable and does not require precise tuning of this parameter.
    
    \item \textbf{Regularization vs. Shot Density ($\lambda_{con}$):} We observe a distinct shift in the need for regularization as the number of shots increases. In the extreme low-shot regime (Aircraft 2-shot), we find a strong positive correlation ($+0.63$) between $\lambda_{con}$ and accuracy, suggesting that heavy regularization is required to prevent overfitting to noisy prototypes. However, as the support set grows (Aircraft 8-shot), this correlation inverts ($-0.34$), indicating that as prototypes become more reliable, the model benefits from relaxing the contrastive constraint.
    
    \item \textbf{Domain-Specific Optimization Landscapes:} Figure~\ref{fig:landscape} illustrates the bivariate optimization landscape ($\gamma$ vs. $\lambda_{con}$) for different domains. While natural image datasets (Flowers, Aircraft) exhibit overlapping high-performance regions, the Eurosat dataset (satellite imagery) displays a distinct landscape topology. This confirms that the optimal balance between the pre-trained prior (CLIP) and the graph adaptation is sensitive to the semantic distance between the target domain and the pre-training data.
\end{enumerate}

\begin{table}[h]
    \centering
    \caption{\textbf{Hyperparameter Optimization Space.} Optimization parameters are tuned on the validation set to adapt to domain granularity and shot availability.}
    \label{tab:search_space}
    \scriptsize
    \begin{tabular}{l c c c}
        \toprule
        \textbf{Hyperparameter} & \textbf{Symbol} & \textbf{Search Range} & \textbf{Scale} \\
        \midrule
        Learning Rate & $\eta$ & $[10^{-5}, 5 \times 10^{-2}]$ & Log-Uniform \\
        Contrastive Weight & $\lambda_{con}$ & $[0.1, 1.0]$ & Uniform \\
        Graph Logit Scale & $\gamma$ & $[0.1, 5.0]$ & Uniform \\
        Focal Loss Gamma & $\gamma_{focal}$ & $[0.5, 3.0]$ & Uniform \\
        Tip-Adapter Alpha & $\alpha$ & $[1.0, 10.0]$ & Uniform \\
        \bottomrule
    \end{tabular}
\end{table}

\section{Ablation of Multi-Scale Patch Hierarchy Granularity}
\label{sec: A7}
We conduct an ablation study on the visual granularity of the Graph Teacher by varying the number and combination of multi-scale patches provided as input nodes (Table \ref{tab:performance_comparison_cvpr}). This analysis validates the optimal configuration for capturing both fine-grained details and global context necessary for robust asymmetric supervision. The results confirm that the performance of the graph-guided approach is sensitive to the diversity of input views, with the established 18-Node MultiScale configuration yielding the best overall performance. This configuration is composed of a heterogeneous set of views: the global image, a $2 \times 2$ grid (4 patches), a $3 \times 3$ grid (9 patches), and vertical/horizontal halves (4 patches). For the Caltech dataset, increasing the node count beyond the optimal $\mathbf{18}$ (e.g., to $21$ or $30$ nodes) leads to marginal performance degradation across all shot settings. The lowest performance is observed when using extremely fine-grained granularity, such as the $30$-node setup (Global $+ 2 \times 2 + 3 \times 3 + 4 \times 4$ grids), which introduces additional visual noise or redundancy in the patches, slightly reducing accuracy in the $1$-shot and $16$-shot regimes.On the EuroSAT dataset, which demands reasoning over complex, heterogeneous scales, the performance variance is more pronounced. While the $\mathbf{18}$-node configuration achieves the highest accuracy in $1$-shot ($67.4\%$), $8$-shot ($84.1\%$), and $16$-shot ($\mathbf{89.4\%}$), the addition of a $4 \times 4$ grid (in the $21$-node configuration) offers competitive performance at $2$-shot and $4$-shot. This suggests that while fine-grained grids are beneficial for distinguishing certain local textures, the inclusion of intermediate-scale splits (halves in the 18-node setup) is more effective for maintaining larger semantic structure, particularly important for complex satellite imagery.This ablation demonstrates that a judicious balance of multi-scale views is crucial, confirming that our $\mathbf{18}$-node MultiScale strategy provides the most comprehensive feature representation for knowledge distillation by avoiding the inherent trade-offs associated with single, fixed-scale grid strategies.

\section{Comparision between Cache and Graph model}
\label{sec: A8}
\begin{figure*}[hbt!]
    \centering
    \includegraphics[width=0.85\linewidth]{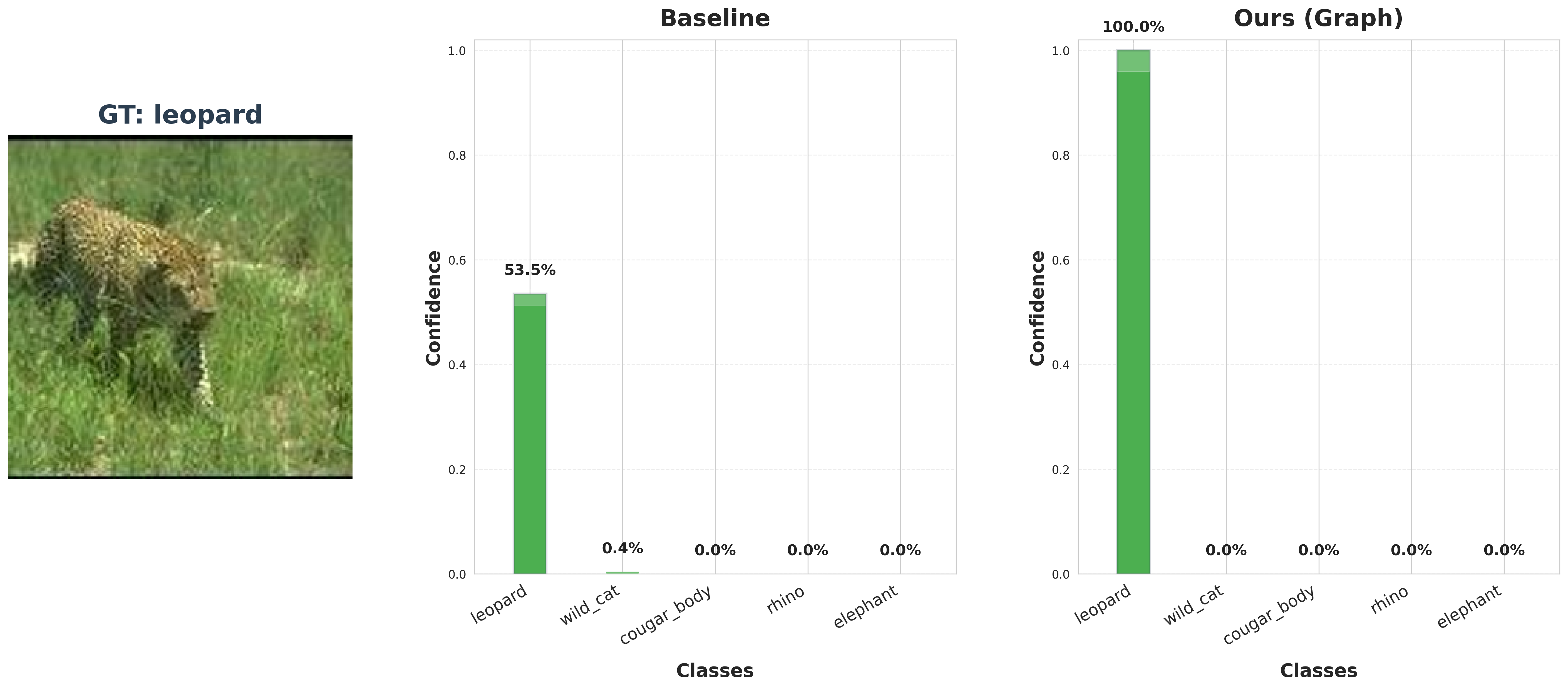}

    \vspace{1em}
    
    \includegraphics[width=0.85\linewidth]{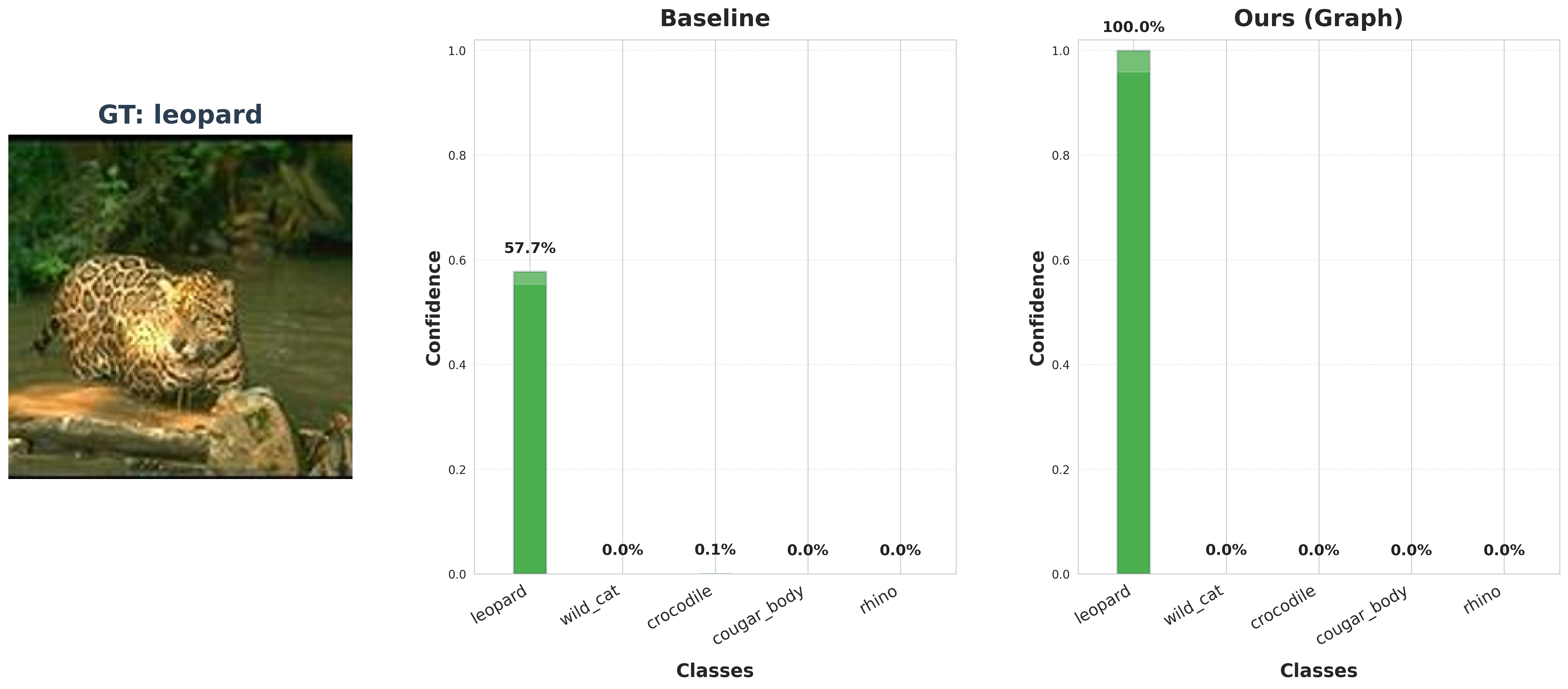}
    
    \vspace{1em} 
    
    \includegraphics[width=0.85\linewidth]{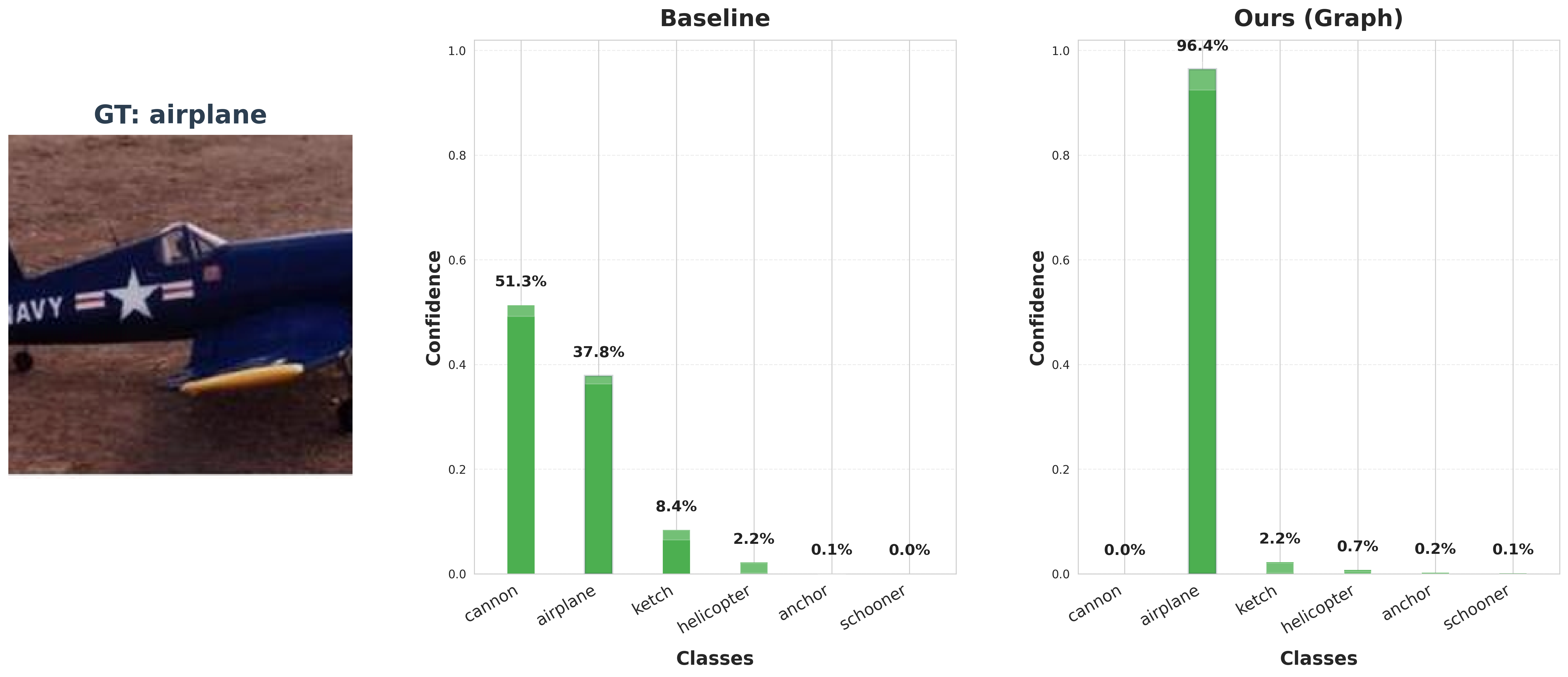}

    \vspace{1em}

    \caption{Quantitative comparison between the Baseline (Tip-Adapter) and Ours (Graph Refined) models on four test examples from the Caltech-101 dataset. Each panel displays the ground truth (GT) label, the input image, and the top-5 classifications with their associated probabilities. Green bars denote the correct class, while red bars indicate an incorrect top-1 prediction. The results demonstrate that the graph-based method corrects the baseline's misclassifications (e.g., in the first and third panels) and significantly improves confidence in correct predictions (e.g., in the second panel).} 
    \label{fig:qualitative_comparison}
\end{figure*}

Quantitative comparisons between the Baseline (Tip-Adapter) and our Graph-Refined model on Caltech-101 are presented in Figure \ref{fig:qualitative_comparison}. Each example includes the ground-truth label, the input image, and the top-5 predicted classes with their corresponding confidence scores. The baseline often exhibits uncertainty or produces incorrect top-1 predictions, while our method consistently yields correct and higher-confidence outputs. In the first and third examples, the baseline misclassifies the input, whereas the graph-refined model correctly identifies the target category with a clear confidence margin. In the second and fourth examples, although the baseline selects the correct class, its probability mass remains distributed across competing categories; our approach suppresses these distractors and concentrates the distribution on the correct label. These results highlight the effectiveness of incorporating inter-class relationships through graph propagation, leading to more stable and discriminative predictions, particularly in challenging or ambiguous cases.

\begin{table}[htbp]
    \centering
    \caption{Few-shot classification accuracy (in \%) for the model across varying numbers of Nodes (rows) and Shots per class (columns) on the Caltech and EuroSAT datasets.}
    \label{tab:performance_comparison_cvpr}
    \begin{tabular}{@{}c || ccccc@{}}
        \toprule
        \textbf{Nodes} & \textbf{1-Shot} & \textbf{2-Shot} & \textbf{4-Shot} & \textbf{8-Shot} & \textbf{16-Shot} \\
        \midrule
        \rowcolor{cyan!10} \multicolumn{6}{l}{\textit{Caltech101}} \\ 
        5 & 94.0 & 94.7 & 95.7 & 95.7 & 96.0 \\
        9 & 94.0 & 94.7 & 95.6 & 95.7 & 96.0 \\
        14 & 94.1 & 94.7 & 95.6 & 95.7 & 96.1 \\
        \textbf{18} & \textbf{94.3} & \textbf{95.0} & \textbf{95.7} & \textbf{95.8} & \textbf{96.3} \\
        21 & 94.1 & 94.8 & 95.6 & 95.7 & 96.2 \\
        30 & 93.8 & 94.9 & 95.6 & 95.6 & 96.0 \\
        \midrule
        
        \rowcolor{cyan!10} \multicolumn{6}{l}{\textit{EuroSAT}} \\ 
        5 & 64.5 & 72.6 & 79.6 & 83.6 & 88.4 \\
        9 & 66.2 & 73.5 & 79.4 & 81.8 & 86.2 \\
        14 & 62.6 & 74.7 & 76.9 & 81.2 & 86.8 \\
        \textbf{18} & \textbf{67.4} & 74.9 & 80.3 & \textbf{84.1} & \textbf{89.4} \\
        21 & 66.8 & \textbf{75.3} & \textbf{81.1} & 83.0 & 88.5 \\
        30 & 65.4 & 72.6 & 77.0 & 83.0 & 88.6 \\
        \bottomrule
    \end{tabular}
\end{table}

\clearpage
\raggedbottom

\section{Additional Ablation Studies}
\label{sec: A9}

\begin{table}[H]
    \centering
    \caption{Impact of Text Nodes in MGT.}
    \label{tab:text_modality_ablation}
    \setlength{\tabcolsep}{3pt} 
    \begin{tabular}{@{}l || c c c c c@{}}
        \toprule
        \textbf{Configuration} & \textbf{1 Shot} & \textbf{2 Shot} & \textbf{4 Shot} & \textbf{8 Shot} & \textbf{16 Shot} \\
        \midrule
        \rowcolor{cyan!10} \multicolumn{6}{l}{\textit{EuroSAT}} \\
        w/o Text & 64.6 & 70.5 & 75.5 & 79.8 & 84.2 \\
        w/ Text & \textbf{67.4} & \textbf{74.9} & \textbf{80.3} & \textbf{84.1} & \textbf{89.4} \\
        \midrule
        \rowcolor{cyan!10} \multicolumn{6}{l}{\textit{Caltech101}} \\
        w/o Text & 93.6 & 93.6 & 95.1 & 95.2 & 95.7 \\
        w/ Text & \textbf{94.3} & \textbf{95.0} & \textbf{95.7} & \textbf{95.8} & \textbf{96.3} \\
        \midrule
        \rowcolor{cyan!10} \multicolumn{6}{l}{\textit{Food101}} \\
        w/o Text & 85.8 & 86.1 & 86.3 & 86.3 & 87.1 \\
        w/ Text & \textbf{86.2} & \textbf{86.5} & \textbf{86.7} & \textbf{86.8} & \textbf{87.5} \\
        \bottomrule
    \end{tabular}
\end{table}
\vspace{-1em}

\begin{table}[htbp]
    \centering
    \caption{Comparison of MGT vs. Transformer architectures across different datasets and shot configurations.}
    \label{tab:mgt_vs_trans}
    \resizebox{\linewidth}{!}{%
    \begin{tabular}{@{}cc || ccccc@{}}
        \toprule
        \textbf{MGT} & \textbf{Trans} & \textbf{1 Shot} & \textbf{2 Shot} & \textbf{4 Shot} & \textbf{8 Shot} & \textbf{16 Shot} \\
        \midrule
        \rowcolor{cyan!10} \multicolumn{7}{l}{\textit{EuroSAT}} \\
        \rmark & \gmark & 62.9 & 72.1 & 77.4 & 80.8 & 87.5 \\
        \gmark & \rmark & \textbf{67.4} & \textbf{74.9} & \textbf{80.3} & \textbf{84.1} & \textbf{89.4} \\
        \midrule
        \rowcolor{cyan!10} \multicolumn{7}{l}{\textit{Aircraft}} \\
        \rmark & \gmark & 27.4 & 31.6 & 34.9 & 41.5 & 45.7 \\
        \gmark & \rmark & \textbf{31.0} & \textbf{34.8} & \textbf{38.3} & \textbf{44.2} & \textbf{48.4} \\
        \midrule
        \rowcolor{cyan!10} \multicolumn{7}{l}{\textit{UCF101}} \\
        \rmark & \gmark & 71.8 & 74.6 & 78.3 & 80.2 & 80.5 \\
        \gmark & \rmark & \textbf{74.5} & \textbf{77.9} & \textbf{81.7} & \textbf{83.9} & \textbf{84.9} \\
        \bottomrule
    \end{tabular}%
    }
\end{table}
\vspace{-1em}

\noindent\textbf{Impact of Text Nodes in Graph Topology.} We evaluate the contribution of text embeddings as structural nodes in Table~\ref{tab:text_modality_ablation}. A baseline (``w/o Text'') constructs a visual-only graph can aggregate patch features but lacks semantic guidance during the reasoning process. This is insufficient for disentangling complex spectral signatures, as seen on EuroSAT. Our proposed method (``w/ Text'') explicitly incorporates text nodes, enabling visual features to attend to and align with high-level semantic concepts layer-by-layer. This ``steers'' the visual representation toward the correct prototype before the final classification stage. The importance of this mechanism is validated by the substantial and widening performance gap at higher shots, indicating that this semantic guidance remains critical even as visual data increases.

\textbf{Ablation on Architecture: MGT vs. Standard Transformer.} As detailed in Table~\ref{tab:mgt_vs_trans}, we evaluate the few-shot generalization capabilities of our proposed MGT architecture against a standard Transformer baseline across three diverse visual domains: remote sensing (EuroSAT), fine-grained object classification (Aircraft), and video action recognition (UCF101). The empirical results demonstrate that the MGT consistently outpaces the standard Transformer across all shot configurations ($K \in \{1, 2, 4, 8, 16\}$). Notably, the MGT architecture exhibits pronounced sample efficiency in highly constrained data regimes, yielding substantial absolute performance gains of +4.5\% on EuroSAT and +3.6\% on FGVC-Aircraft under the extreme 1-shot setting. This consistent performance advantage across distinct modalities and varying data availability highlights the superior feature representation and robustness of the MGT framework when compared to traditional Transformer counterparts.

%% file: ArXiv.bbl
\begin{thebibliography}{48}
\providecommand{\natexlab}[1]{#1}
\providecommand{\url}[1]{\texttt{#1}}
\expandafter\ifx\csname urlstyle\endcsname\relax
  \providecommand{\doi}[1]{doi: #1}\else
  \providecommand{\doi}{doi: \begingroup \urlstyle{rm}\Url}\fi

\bibitem[Bera et~al.(2022)Bera, Wharton, Liu, Bessis, and Behera]{bera2022srgnn}
Asish Bera, Zachary Wharton, Yonghuai Liu, Nik Bessis, and Ardhendu Behera.
\newblock Sr-gnn: Spatial relation-aware graph neural network for fine-grained image categorization.
\newblock \emph{IEEE Transactions on Image Processing}, 31:\penalty0 6017--6031, 2022.

\bibitem[Bossard et~al.(2014)Bossard, Guillaumin, and Van~Gool]{bossard2014food}
Lukas Bossard, Matthieu Guillaumin, and Luc Van~Gool.
\newblock Food-101--mining discriminative components with random forests.
\newblock In \emph{European conference on computer vision}, pages 446--461. Springer, 2014.

\bibitem[Chen et~al.(2023)Chen, Yao, Song, Li, Rao, and Zhang]{chen2022prompt}
Guangyi Chen, Weiran Yao, Xiangchen Song, Xinyue Li, Yongming Rao, and Kun Zhang.
\newblock Prompt learning with optimal transport for vision-language models.
\newblock In \emph{Proceedings of the ICLR}, 2023.

\bibitem[Cimpoi et~al.(2014)Cimpoi, Maji, Kokkinos, Mohamed, and Vedaldi]{cimpoi2014describing}
Mircea Cimpoi, Subhransu Maji, Iasonas Kokkinos, Sammy Mohamed, and Andrea Vedaldi.
\newblock Describing textures in the wild.
\newblock In \emph{Proceedings of the IEEE conference on computer vision and pattern recognition}, pages 3606--3613, 2014.

\bibitem[Deng et~al.(2009)Deng, Dong, Socher, Li, Li, and Fei-Fei]{deng2009imagenet}
Jia Deng, Wei Dong, Richard Socher, Li-Jia Li, Kai Li, and Li Fei-Fei.
\newblock Imagenet: A large-scale hierarchical image database.
\newblock In \emph{2009 IEEE conference on computer vision and pattern recognition}, pages 248--255. Ieee, 2009.

\bibitem[Farina et~al.(2025)Farina, Mancini, Iacca, and Ricci]{farina2025rethinking}
Matteo Farina, Massimiliano Mancini, Giovanni Iacca, and Elisa Ricci.
\newblock Rethinking few-shot adaptation of vision-language models in two stages.
\newblock In \emph{Proceedings of the Computer Vision and Pattern Recognition Conference}, pages 29989--29998, 2025.

\bibitem[Fei-Fei et~al.(2004)Fei-Fei, Fergus, and Perona]{fei2004learning}
Li Fei-Fei, Rob Fergus, and Pietro Perona.
\newblock Learning generative visual models from few training examples: An incremental bayesian approach tested on 101 object categories.
\newblock In \emph{2004 conference on computer vision and pattern recognition workshop}, pages 178--178. IEEE, 2004.

\bibitem[Fu et~al.(2017)Fu, Zheng, and Mei]{fu2017look}
Jianlong Fu, Heliang Zheng, and Tao Mei.
\newblock Look closer to see better: Recurrent attention convolutional neural network for fine-grained image recognition.
\newblock In \emph{Proceedings of the IEEE conference on computer vision and pattern recognition}, pages 4438--4446, 2017.

\bibitem[Gao et~al.(2024)Gao, Geng, Zhang, Ma, Fang, Zhang, Li, and Qiao]{gao2024clipadapter}
Peng Gao, Shijie Geng, Renrui Zhang, Teli Ma, Rongyao Fang, Yongfeng Zhang, Hongsheng Li, and Yu Qiao.
\newblock Clip-adapter: Better vision-language models with feature adapters.
\newblock \emph{International Journal of Computer Vision}, 132\penalty0 (2):\penalty0 581--595, 2024.

\bibitem[He et~al.(2022)He, Chen, Liu, Kortylewski, Yang, Bai, and Wang]{he2022transfg}
Ju He, Jie-Neng Chen, Shuai Liu, Adam Kortylewski, Cheng Yang, Yutong Bai, and Changhu Wang.
\newblock Transfg: A transformer architecture for fine-grained recognition.
\newblock In \emph{Proceedings of the AAAI conference on artificial intelligence}, pages 852--860, 2022.

\bibitem[Helber et~al.(2019)Helber, Bischke, Dengel, and Borth]{helber2019eurosat}
Patrick Helber, Benjamin Bischke, Andreas Dengel, and Damian Borth.
\newblock Eurosat: A novel dataset and deep learning benchmark for land use and land cover classification.
\newblock \emph{IEEE Journal of Selected Topics in Applied Earth Observations and Remote Sensing}, 12\penalty0 (7):\penalty0 2217--2226, 2019.

\bibitem[Houlsby et~al.(2019)Houlsby, Giurgiu, Jastrzebski, Morrone, De~Laroussilhe, Gesmundo, Attariyan, and Gelly]{houlsby2019parameter}
Neil Houlsby, Andrei Giurgiu, Stanislaw Jastrzebski, Bruna Morrone, Quentin De~Laroussilhe, Andrea Gesmundo, Mona Attariyan, and Sylvain Gelly.
\newblock Parameter-efficient transfer learning for nlp.
\newblock In \emph{International conference on machine learning}, pages 2790--2799. PMLR, 2019.

\bibitem[Hu et~al.(2022)Hu, Shen, Wallis, Allen-Zhu, Li, Wang, Wang, Chen, et~al.]{hu2022lora}
Edward~J Hu, Yelong Shen, Phillip Wallis, Zeyuan Allen-Zhu, Yuanzhi Li, Shean Wang, Lu Wang, Weizhu Chen, et~al.
\newblock Lora: Low-rank adaptation of large language models.
\newblock \emph{ICLR}, 1\penalty0 (2):\penalty0 3, 2022.

\bibitem[Hu et~al.(2020)Hu, Dong, Wang, and Sun]{hu2020heterogeneous}
Ziniu Hu, Yuxiao Dong, Kuansan Wang, and Yizhou Sun.
\newblock Heterogeneous graph transformer.
\newblock In \emph{Proceedings of the web conference 2020}, pages 2704--2710, 2020.

\bibitem[Huang et~al.(2025)Huang, Jiang, Jiang, Li, Khan, and Wang]{huang2025cosmic}
Fanding Huang, Jingyan Jiang, Qinting Jiang, Hebei Li, Faisal~Nadeem Khan, and Zhi Wang.
\newblock Cosmic: Clique-oriented semantic multi-space integration for robust clip test-time adaptation.
\newblock In \emph{Proceedings of the Computer Vision and Pattern Recognition Conference}, pages 9772--9781, 2025.

\bibitem[Jia et~al.(2021)Jia, Yang, Xia, Chen, Parekh, Pham, Le, Sung, Li, and Duerig]{jia2021scaling}
Chao Jia, Yinfei Yang, Ye Xia, Yi-Ting Chen, Zarana Parekh, Hieu Pham, Quoc Le, Yun-Hsuan Sung, Zhen Li, and Tom Duerig.
\newblock Scaling up visual and vision-language representation learning with noisy text supervision.
\newblock In \emph{International conference on machine learning}, pages 4904--4916. PMLR, 2021.

\bibitem[Jia et~al.(2022)Jia, Tang, Chen, Cardie, Belongie, Hariharan, and Lim]{jia2022visual}
Menglin Jia, Luming Tang, Bor-Chun Chen, Claire Cardie, Serge Belongie, Bharath Hariharan, and Ser-Nam Lim.
\newblock Visual prompt tuning.
\newblock In \emph{European conference on computer vision}, pages 709--727. Springer, 2022.

\bibitem[Jiang et~al.(2025)Jiang, Zhang, Liu, and Shi]{jiang2025causal}
Tianjiao Jiang, Zhen Zhang, Yuhang Liu, and Javen~Qinfeng Shi.
\newblock Causal disentanglement and cross-modal alignment for enhanced few-shot learning.
\newblock In \emph{Proceedings of the IEEE/CVF International Conference on Computer Vision}, pages 890--900, 2025.

\bibitem[Khattak et~al.(2023)Khattak, Rasheed, Maaz, Khan, and Khan]{khattak2023maple}
Muhammad~Uzair Khattak, Hanoona Rasheed, Muhammad Maaz, Salman Khan, and Fahad~Shahbaz Khan.
\newblock Maple: Multi-modal prompt learning.
\newblock In \emph{Proceedings of the IEEE/CVF conference on computer vision and pattern recognition}, pages 19113--19122, 2023.

\bibitem[Krause et~al.(2013)Krause, Stark, Deng, and Fei-Fei]{krause20133d}
Jonathan Krause, Michael Stark, Jia Deng, and Li Fei-Fei.
\newblock 3d object representations for fine-grained categorization.
\newblock In \emph{Proceedings of the IEEE international conference on computer vision workshops}, pages 554--561, 2013.

\bibitem[Li et~al.(2024)Li, Zhang, Lin, Chen, and He]{li2024pixels}
Rongjie Li, Songyang Zhang, Dahua Lin, Kai Chen, and Xuming He.
\newblock From pixels to graphs: Open-vocabulary scene graph generation with vision-language models.
\newblock In \emph{Proceedings of the IEEE/CVF conference on computer vision and pattern recognition}, pages 28076--28086, 2024.

\bibitem[Li et~al.(2023)Li, Lian, Lu, Bai, Chen, and Wang]{li2023graphadapter}
Xin Li, Dongze Lian, Zhihe Lu, Jiawang Bai, Zhibo Chen, and Xinchao Wang.
\newblock Graphadapter: Tuning vision-language models with dual knowledge graph.
\newblock \emph{Advances in Neural Information Processing Systems}, 36:\penalty0 13448--13466, 2023.

\bibitem[Lin et~al.(2017)Lin, Goyal, Girshick, He, and Doll{\'a}r]{lin2017focal}
Tsung-Yi Lin, Priya Goyal, Ross Girshick, Kaiming He, and Piotr Doll{\'a}r.
\newblock Focal loss for dense object detection.
\newblock In \emph{Proceedings of the IEEE international conference on computer vision}, pages 2980--2988, 2017.

\bibitem[Maji et~al.(2013)Maji, Rahtu, Kannala, Blaschko, and Vedaldi]{maji2013fineaircraft}
Subhransu Maji, Esa Rahtu, Juho Kannala, Matthew Blaschko, and Andrea Vedaldi.
\newblock Fine-grained visual classification of aircraft.
\newblock \emph{arXiv preprint arXiv:1306.5151}, 2013.

\bibitem[Nilsback and Zisserman(2008)]{nilsback2008automatedflowers}
Maria-Elena Nilsback and Andrew Zisserman.
\newblock Automated flower classification over a large number of classes.
\newblock In \emph{2008 Sixth Indian conference on computer vision, graphics \& image processing}, pages 722--729. IEEE, 2008.

\bibitem[Parkhi et~al.(2012)Parkhi, Vedaldi, Zisserman, and Jawahar]{parkhi2012cats}
Omkar~M Parkhi, Andrea Vedaldi, Andrew Zisserman, and CV Jawahar.
\newblock Cats and dogs.
\newblock In \emph{2012 IEEE conference on computer vision and pattern recognition}, pages 3498--3505. IEEE, 2012.

\bibitem[Radford et~al.(2021)Radford, Kim, Hallacy, Ramesh, Goh, Agarwal, Sastry, Askell, Mishkin, Clark, et~al.]{radford2021clip}
Alec Radford, Jong~Wook Kim, Chris Hallacy, Aditya Ramesh, Gabriel Goh, Sandhini Agarwal, Girish Sastry, Amanda Askell, Pamela Mishkin, Jack Clark, et~al.
\newblock Learning transferable visual models from natural language supervision.
\newblock In \emph{International conference on machine learning}, pages 8748--8763. PmLR, 2021.

\bibitem[Ren et~al.(2025)Ren, Chen, Wang, and Hua]{ren2025vpt}
Li Ren, Chen Chen, Liqiang Wang, and Kien Hua.
\newblock Da-vpt: Semantic-guided visual prompt tuning for vision transformers.
\newblock In \emph{Proceedings of the Computer Vision and Pattern Recognition Conference}, pages 4353--4363, 2025.

\bibitem[Sikdar et~al.(2025)Sikdar, Liu, Kedarisetty, Zhao, Ahmed, and Behera]{sikdar2025interweaving}
Arindam Sikdar, Yonghuai Liu, Siddhardha Kedarisetty, Yitian Zhao, Amr Ahmed, and Ardhendu Behera.
\newblock Interweaving insights: High-order feature interaction for fine-grained visual recognition.
\newblock \emph{International Journal of Computer Vision}, 133\penalty0 (4):\penalty0 1755--1779, 2025.

\bibitem[Silva-Rodriguez et~al.(2024)Silva-Rodriguez, Hajimiri, Ben~Ayed, and Dolz]{silva2024closer}
Julio Silva-Rodriguez, Sina Hajimiri, Ismail Ben~Ayed, and Jose Dolz.
\newblock A closer look at the few-shot adaptation of large vision-language models.
\newblock In \emph{Proceedings of the IEEE/CVF Conference on Computer Vision and Pattern Recognition}, pages 23681--23690, 2024.

\bibitem[Soomro et~al.(2012)Soomro, Zamir, and Shah]{soomro2012ucf101}
Khurram Soomro, Amir~Roshan Zamir, and Mubarak Shah.
\newblock Ucf101: A dataset of 101 human actions classes from videos in the wild.
\newblock \emph{arXiv preprint arXiv:1212.0402}, 2012.

\bibitem[Vaswani et~al.(2017)Vaswani, Shazeer, Parmar, Uszkoreit, Jones, Gomez, Kaiser, and Polosukhin]{vaswani2017attention}
Ashish Vaswani, Noam Shazeer, Niki Parmar, Jakob Uszkoreit, Llion Jones, Aidan~N Gomez, {\L}ukasz Kaiser, and Illia Polosukhin.
\newblock Attention is all you need.
\newblock \emph{Advances in neural information processing systems}, 30, 2017.

\bibitem[Wah et~al.(2011)Wah, Branson, Welinder, Perona, and Belongie]{wah2011caltech}
Catherine Wah, Steve Branson, Peter Welinder, Pietro Perona, and Serge Belongie.
\newblock The caltech-ucsd birds-200-2011 dataset.
\newblock 2011.

\bibitem[Wu et~al.(2024)Wu, Li, Chen, Yu, Gu, and Wang]{wu20243d}
Zizhao Wu, Haohan Li, Gongyi Chen, Zhou Yu, Xiaoling Gu, and Yigang Wang.
\newblock 3d question answering with scene graph reasoning.
\newblock In \emph{Proceedings of the 32nd ACM International Conference on Multimedia}, pages 1370--1378, 2024.

\bibitem[Xiao et~al.(2010)Xiao, Hays, Ehinger, Oliva, and Torralba]{xiao2010sun}
Jianxiong Xiao, James Hays, Krista~A Ehinger, Aude Oliva, and Antonio Torralba.
\newblock Sun database: Large-scale scene recognition from abbey to zoo.
\newblock In \emph{2010 IEEE computer society conference on computer vision and pattern recognition}, pages 3485--3492. IEEE, 2010.

\bibitem[Yang et~al.(2024)Yang, An, Huang, Bi, Yu, Yang, Diao, and Xu]{yang2024clip}
Chuanguang Yang, Zhulin An, Libo Huang, Junyu Bi, Xinqiang Yu, Han Yang, Boyu Diao, and Yongjun Xu.
\newblock Clip-kd: An empirical study of clip model distillation.
\newblock In \emph{Proceedings of the IEEE/CVF Conference on Computer Vision and Pattern Recognition}, pages 15952--15962, 2024.

\bibitem[Yao et~al.(2023)Yao, Zhang, and Xu]{yao2023visual}
Hantao Yao, Rui Zhang, and Changsheng Xu.
\newblock Visual-language prompt tuning with knowledge-guided context optimization.
\newblock In \emph{Proceedings of the IEEE/CVF conference on computer vision and pattern recognition}, pages 6757--6767, 2023.

\bibitem[Yu et~al.(2023)Yu, Lu, Jin, Chen, and Wang]{yu2023task}
Tao Yu, Zhihe Lu, Xin Jin, Zhibo Chen, and Xinchao Wang.
\newblock Task residual for tuning vision-language models.
\newblock In \emph{Proceedings of the IEEE/CVF conference on computer vision and pattern recognition}, pages 10899--10909, 2023.

\bibitem[Zanella and Ben~Ayed(2024)]{zanella2024low}
Maxime Zanella and Ismail Ben~Ayed.
\newblock Low-rank few-shot adaptation of vision-language models.
\newblock In \emph{Proceedings of the IEEE/CVF Conference on Computer Vision and Pattern Recognition}, pages 1593--1603, 2024.

\bibitem[Zanella et~al.(2024)Zanella, G{\'e}rin, and Ayed]{zanella2024boosting}
Maxime Zanella, Beno{\^\i}t G{\'e}rin, and Ismail Ayed.
\newblock Boosting vision-language models with transduction.
\newblock \emph{Advances in Neural Information Processing Systems}, 37:\penalty0 62223--62256, 2024.

\bibitem[Zeng et~al.(2024)Zeng, Han, Wang, Wu, Geng, Huangg, Wu, and Liu]{zeng2024visual}
Runjia Zeng, Cheng Han, Qifan Wang, Chunshu Wu, Tong Geng, Lifu Huangg, Ying~Nian Wu, and Dongfang Liu.
\newblock Visual fourier prompt tuning.
\newblock \emph{Advances in Neural Information Processing Systems}, 37:\penalty0 5552--5585, 2024.

\bibitem[Zhai et~al.(2022)Zhai, Wang, Mustafa, Steiner, Keysers, Kolesnikov, and Beyer]{zhai2022lit}
Xiaohua Zhai, Xiao Wang, Basil Mustafa, Andreas Steiner, Daniel Keysers, Alexander Kolesnikov, and Lucas Beyer.
\newblock Lit: Zero-shot transfer with locked-image text tuning.
\newblock In \emph{Proceedings of the IEEE/CVF conference on computer vision and pattern recognition}, pages 18123--18133, 2022.

\bibitem[Zhang et~al.(2022)Zhang, Zhang, Fang, Gao, Li, Dai, Qiao, and Li]{tipadapter}
Renrui Zhang, Wei Zhang, Rongyao Fang, Peng Gao, Kunchang Li, Jifeng Dai, Yu Qiao, and Hongsheng Li.
\newblock Tip-adapter: Training-free adaption of clip for few-shot classification.
\newblock In \emph{European conference on computer vision}, pages 493--510. Springer, 2022.

\bibitem[Zhang et~al.(2023)Zhang, Hu, Li, Huang, Deng, Qiao, Gao, and Li]{zhang2023prompt}
Renrui Zhang, Xiangfei Hu, Bohao Li, Siyuan Huang, Hanqiu Deng, Yu Qiao, Peng Gao, and Hongsheng Li.
\newblock Prompt, generate, then cache: Cascade of foundation models makes strong few-shot learners.
\newblock In \emph{Proceedings of the IEEE/CVF conference on computer vision and pattern recognition}, pages 15211--15222, 2023.

\bibitem[Zhang et~al.(2024)Zhang, Zhang, Deng, Zhang, Lin, Huang, Zhang, and Yu]{zhang2024ta}
Wenbo Zhang, Yifan Zhang, Yuyang Deng, Wenlong Zhang, Jianfeng Lin, Binqiang Huang, Jinlu Zhang, and Wenhao Yu.
\newblock Ta-adapter: Enhancing few-shot clip with task-aware encoders.
\newblock \emph{Pattern Recognition}, 153:\penalty0 110559, 2024.

\bibitem[Zhou et~al.(2022{\natexlab{a}})Zhou, Yang, Loy, and Liu]{zhou2022conditional}
Kaiyang Zhou, Jingkang Yang, Chen~Change Loy, and Ziwei Liu.
\newblock Conditional prompt learning for vision-language models.
\newblock In \emph{Proceedings of the IEEE/CVF conference on computer vision and pattern recognition}, pages 16816--16825, 2022{\natexlab{a}}.

\bibitem[Zhou et~al.(2022{\natexlab{b}})Zhou, Yang, Loy, and Liu]{zhou2022learning}
Kaiyang Zhou, Jingkang Yang, Chen~Change Loy, and Ziwei Liu.
\newblock Learning to prompt for vision-language models.
\newblock \emph{International Journal of Computer Vision}, 130\penalty0 (9):\penalty0 2337--2348, 2022{\natexlab{b}}.

\bibitem[Zhu et~al.(2023)Zhu, Niu, Han, Wu, and Zhang]{zhu2023prompt}
Beier Zhu, Yulei Niu, Yucheng Han, Yue Wu, and Hanwang Zhang.
\newblock Prompt-aligned gradient for prompt tuning.
\newblock In \emph{Proceedings of the IEEE/CVF international conference on computer vision}, pages 15659--15669, 2023.

\end{thebibliography}
